\documentclass[sigconf,nonacm]{acmart}
\usepackage{subcaption} % in the preamble
\AtBeginDocument{%
  }

\copyrightyear{2026}
\acmYear{2026}
\setcopyright{cc}
\setcctype{by}
\acmConference[CIKM '26] {Proceedings of the 35th ACM International Conference on Information and Knowledge Management}{November 7--11, 2026}{Rome, Italy.}
\acmBooktitle{Proceedings of the 35th ACM International Conference on Information and Knowledge Management (CIKM '26), November 7--11, 2026, Rome, Italy}
\acmISBN{979-8-4007-2539-5/2026/11}
\acmDOI{10.1145/3799682.3841045}

\begin{document}

\title{Feature Evolution and Migration during Vision Transformer Training}
\titlenote{This manuscript is an extended version of the paper published
in the Proceedings of the 35th ACM International Conference on Information
and Knowledge Management (CIKM 2026). The conference version is available
at \url{https://doi.org/10.1145/3799682.3841045}.}
\author{Joonas Järve}

\affiliation{%
 \department{Institute of Computer Science}
  \institution{University of Tartu}
  \city{Tartu}
  \country{Estonia}}
%%\authornote{Both authors contributed equally to this research.}
\email{joonas.jarve@ut.ee}

\author{Halil Ibrahim Aysel}
\affiliation{%
\department{Institute of Computer Science}
  \institution{University of Tartu}
  \city{Tartu}
  \country{Estonia}}
\email{halil.ibrahim.aysel@ut.ee}

\author{Tarun Khajuria}

\affiliation{%
\department{Institute of Computer Science}
  \institution{University of Tartu}
  \city{Tartu}
  \country{Estonia}}
\email{tarun.khajuria@ut.ee}

\author{Meelis Kull}

\affiliation{%
\department{Institute of Computer Science}
  \institution{University of Tartu}
  \city{Tartu}
  \country{Estonia}}
\email{meelis.kull@ut.ee}

%%\renewcommand{\shortauthors}{Trovato et al.}

\begin{abstract}
 We present a novel view on feature evolution in Vision Transformers (ViTs) by visualizing the training process over two dimensions -- network depth (layer) and training time (epochs). We employ Sparse Autoencoders (SAEs) to extract candidate sparse features from CLS-token representations and compare their activation profiles across epoch--layer pairs. This allows us to study feature-level dynamics that are not directly visible from representation-level similarity measures. Furthermore, we demonstrate how this framework of feature evolution allows us to describe \textit{feature migration}, the change in the layer where a feature is most detectable during training. Our experiments show that migration is concentrated early in training, occurs more often toward earlier layers than toward deeper layers, and declines as feature organization stabilizes. We further find that deeper layers stabilize earlier and more strongly than shallow layers. The results show that our approach can be employed as a tool for understanding how ViTs learn and evolve.
 
\end{abstract}




%%\received{20 February 2007}
%%\received[revised]{12 March 2009}
%%\received[accepted]{5 June 2009}

\maketitle

\section{Introduction}
  
Vision Transformers (ViTs) \citep{dosovitskiy2020vit} have become standard backbones for visual recognition and multimodal learning, but we still have limited understanding of how their internal visual features are formed during training. Prior work has shown that both convolutional neural networks (CNNs) \citep{lecun1989backpropagation} and ViTs exhibit broad hierarchical structure across depth, progressing from simpler patterns in early layers toward more semantic structure in deeper layers \citep{huang2021semantic,ghiasi2022vision,dorszewski2025colors,fel2024understanding,kim2025interpreting}. Less is known about how it develops: when do visual features first appear, how long do they persist, and do they remain in the same layers throughout training?
\begin{figure}[ht]
  \centering
  \includegraphics[width=\linewidth,trim={0 0cm 1.5cm 0},clip]{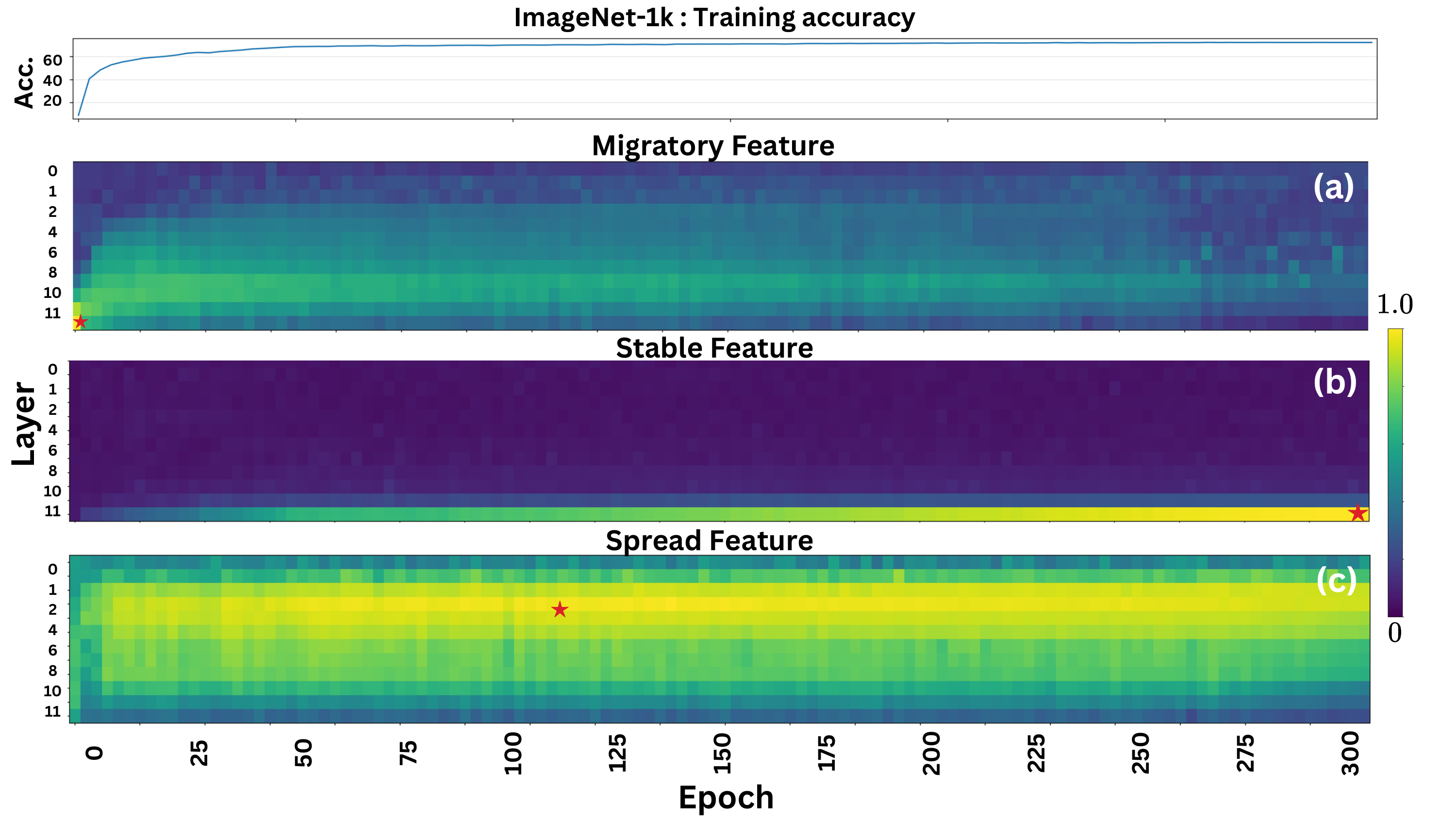}
  \caption{Examples of distinct feature-evolution types observed during ViT training, shown in an epoch-by-layer view. {\color{red}$\star$} represents the reference feature.}\label{fig:main_fig}
\end{figure}
Existing analysis tools operate either at the level of individual neurons, which are often entangled, or at the level of whole representations, where scalar similarity scores can obscure which features are actually shared between representations. In particular, alignment measures such as CKA \citep{kornblith2019similarity} are informative about representation-level similarity, but do not reveal whether two layers or checkpoints contain the same features \citep{vielhaben2024beyond}, rendering feature evolution tracking difficult. Linear probes \citep{alain2016understanding} can track predefined human-labeled attributes, but only for attributes that are already annotated. As \citet{yun2021transformer} put it: \emph{``Probing tasks can only verify whether a certain prior structure is learned in a language model. They can not reveal the structures beyond our prior knowledge.''} We therefore need feature-level methods that can track both known human-interpretable attributes and features discovered directly from model activations.

In this work, we study ViT training along two axes: time, represented by training checkpoints, and depth, represented by network layers. As a growing body of literature suggests that sparse autoencoders (SAEs) \citep{huben2024sparse,ng2011sparse} can recover more disentangled and often more monosemantic features than raw hidden units, partly mitigating the effects of superposition \citep{huben2024sparse,bricken2023towards}, we base our feature discovery on SAEs. At each checkpoint and layer, we analyze the model’s representations using SAEs to extract sparse latent features from the model’s activations. Throughout the paper, we use the term \emph{feature} broadly: some features are human-interpretable, while others may be model-internal factors \citep{till_2024_saes_true_features} without a simple semantic label. This distinction is particularly important in vision, where meaningful internal factors need not align with natural-language categories \citep{karvonen2024evaluating}.

This two-dimensional view allows us to study feature migration. We say that a feature migrates when the layer where it is most detectable changes during training. For example, a feature that is initially visible only in deeper layers may later become detectable in earlier layers (example \textbf{a} in Figure \ref{fig:main_fig}), suggesting that the model has reorganized where that information is represented. Conversely, some features may remain local and become increasingly stable as training progresses (example \textbf{b} in Figure \ref{fig:main_fig}) and some spread across layers (example \textbf{c} in Figure \ref{fig:main_fig}).

Our experiments use this framework to ask three questions. First, how do visual features emerge and stabilize across ViT layers during training? Second, do features migrate from deeper to earlier layers, or from earlier to deeper layers? Third, how do these dynamics depend on the dataset scale? We find that feature migration is most pronounced early in training, that deeper-layer features become more stable over time, and dataset scale does not influence most of the migration metrics.

Our contributions are as follows:
\begin{enumerate}
    \item We introduce an epoch-by-layer framework for tracking visual features in ViTs during training.
    %\item We find that probing and SAE feature evolution patterns cluster similarly via the feature migration metric.
    \item We define and measure feature migration, showing how features shift across layers as training progresses.
    \item We provide empirical evidence that ViT feature organization stabilizes over training, with much of the observed migration occurring in the early stages.
\end{enumerate}

The code of this work is available on GitHub \footnote{\url{https://github.com/joonasrooben/VIT-Feature-Evolution-and-Migration}}.
\section{Related Work}
\label{sec:related_work}

\paragraph{Feature-level interpretation in vision models.} A large body of work studies what internal representations of vision models encode. Early visualization methods inspected highly activating image patches \citep{liu2016towards}, while Network Dissection \citep{bau2017network} quantified the alignment between hidden units and human-interpretable features, also known as concepts. Other concept-based methods, such as TCAV  \citep{kim2018interpretability} and concept bottleneck models \citep{koh2020concept}, analyze user-defined concepts or enforce concept-level intermediate representations. In parallel, studies of CNNs and ViTs have shown that visual representations often become more abstract with depth, moving from colors, textures, and local patterns toward object- or class-level structure \citep{huang2021semantic,geirhos2018imagenet,bau2017network,ghiasi2022vision,dorszewski2025colors,fel2024understanding,kim2025interpreting,lim2024sparse,olson2025probing}. These works motivate feature-level analysis, but they usually study trained models or individual layers rather than how features evolve throughout training. %What remains less understood is whether this hierarchy itself changes over training: for example, whether features that first appear in deeper layers later shift toward earlier layers and vice versa.

\paragraph{Representation similarity and training dynamics.} Another line of work compares internal representations across layers, checkpoints, datasets, or architectures using measures such as SVCCA \citep{raghu2017svcca} and CKA \citep{kornblith2019similarity}. These methods are useful for measuring global similarity between activation spaces \citep{kornblith2019similarity, huh2024position,ciernik2024objective} and have revealed important differences between CNNs and ViTs, including the role of residual connections and early global information flow in ViTs  \citep{raghu2021vision,fel2024understanding}. However, global representational similarity does not identify which features are shared \citep{vielhaben2024beyond, kondapaneni2025representational,kondapaneni2025representational_nips}. Two representations may be similar while containing different features, and two representations may support related features even if their activation spaces are not globally aligned. Our work therefore complements representation-level analysis with feature-level tracking.

\paragraph{Sparse feature discovery.} Sparse autoencoders and related dictionary learning methods decompose hidden activations into sparse latent variables \citep{zheng2021deep}. In vision models, recent works \citep{lim2024sparse,olson2025probing,pach2025sparse,stevens2025interpretable} have shown that such latents can correspond to meaningful visual patterns in CLIP \citep{radford2021learning}, DINOv2 \citep{oquab2023dinov2}, and other ViT-based encoders. These methods are attractive because they do not require a predefined vocabulary of human concepts. At the same time, SAE latents should not be treated as guaranteed atomic or human-interpretable concepts \citep{karvonen2025saebench,leask2025sparse,peng2025use}: they may be unstable across runs, split across related features, or represent mixtures \citep{marks2024enhancing,chanin2024absorption,fel2025archetypal}. We therefore treat SAE latents as candidate sparse features and evaluate their continuity through their activation profiles. Overall, studies establish SAEs as a promising tool for vision interpretability, but principally our epoch-by-layer view is not limited to the method used for feature discovery; any other method could be used as well (e.g., NMF \citep{lee1999learning}, PCA, or K-SVD \citep{valentin2025db}).

\paragraph{Feature evolution and migration.} The closest work to ours studies how representations or sparse features change during training. Prior studies have analyzed training dynamics in CNNs and ViTs \citep{sharon2024does, kim2025interpreting,fel2024understanding}, and recent SAE-based work has tracked feature dynamics in language models \citep{xu2024tracking}. However, existing work does not provide an epoch-by-layer account of feature evolution in ViTs. In this paper, we study whether visual features emerge, persist, disappear, or migrate between layers during training. This gives a feature-level view of ViT learning that is not captured by either static interpretability methods or scalar representation-similarity measures. Since state-of-the-art visual backbones such as DINOv2 and CLIP-style encoders are built on the ViT architecture, understanding their feature-level training dynamics is therefore crucial.

Prior work has interpreted features, compared representations, and studied training dynamics, but to the best of our knowledge, has not tracked feature-level migration across both ViT depth and training time.

\section{Methodology}
In this section, we introduce our method for computing similarities between features when the features are not predefined with labels specifying for each image whether it has this property or not. When features are predefined, one can simply use linear probing and track the features in our proposed epoch-by-layer map.

First, we show how to use SAEs to extract candidate sparse features from CLS-token representations, and second, explain how to compare their activation profiles across epoch--layer pairs. This allows us to study feature-level dynamics that are not directly visible from representation-level similarity measures.

\begin{figure*}[ht]
  \centering
  \includegraphics[width=\linewidth,trim={0 2.6cm 0 0},clip]{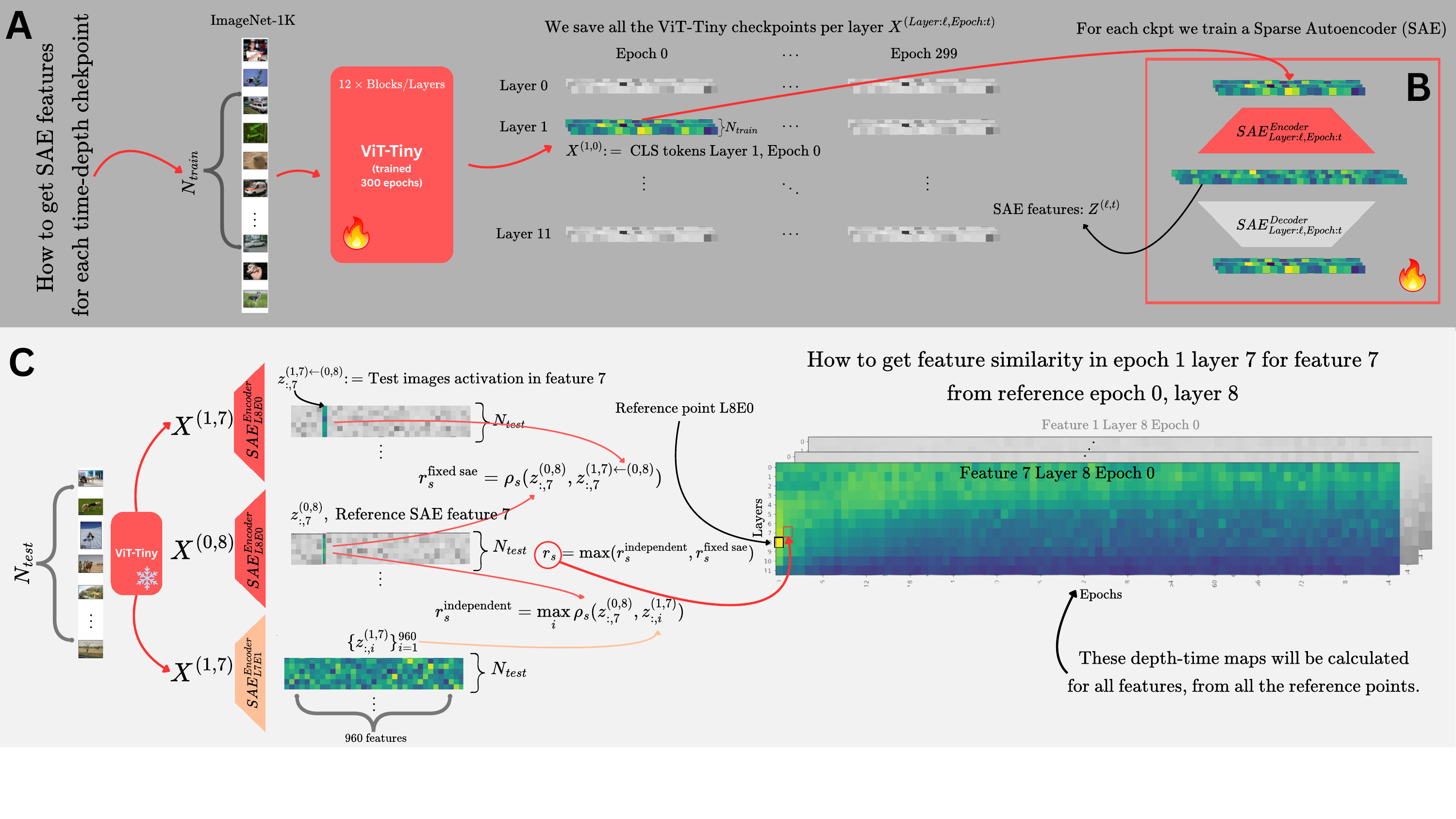} 
  \caption{Overview of the methodology: (A) extraction of image representations from every ViT checkpoint–layer pair; (B) SAE training at each pair; and (C) computation of feature-similarity scores for one example.}\label{fig:method}
\end{figure*}

Let $f_{\theta^{(t)}}$ denote a ViT backbone at training checkpoint $t$ and let $h^{(t,\ell)}(a) \in \mathbb{R}^{d_\ell}$ denote its CLS-token representation at layer $\ell$ for input $a$. Given an image dataset $\mathcal{D}=\{a_i\}_{i=1}^n$, we form a representation matrix
\begin{equation}
X^{(t,\ell)} = 
\begin{bmatrix}
x_1 = h^{(t,\ell)}(a_1)^\top\\
\vdots\\
x_n = h^{(t,\ell)}(a_n)^\top
\end{bmatrix}
\in \mathbb{R}^{n \times d_\ell}.
\end{equation}
This representation extraction phase is illustrated in part \textbf{A} of Figure \ref{fig:method}, showing that representations are extracted for each epoch-layer pair.

%\subsection{Sparse Autoencoder-based approach}
%\label{subsec:sae}
Now, we go over our SAE setup and explain how we get SAE features. This is illustrated in part \textbf{B} of Figure \ref{fig:method}.
Given activations $x \in X^{(t,\ell)}$ from a fixed layer and checkpoint, a sparse autoencoder (SAE) maps $x$ to a sparse latent code $z \in \mathbb{R}^{m}$ and reconstructs it as
\begin{align}
z &= \mathrm{ReLU}(W_{enc} (x - b_{dec}) + b_{enc}), \\
\hat{x} &= W_{dec} z + b_{dec},
\end{align}
where $W_{enc} \in \mathbb{R}^{m \times d}$ and $W_{dec} \in \mathbb{R}^{d \times m}$ are the encoder and decoder weights; $b_{enc}$ and $b_{dec}$ are the encoder and decoder  biases, respectively \citep{bricken2023towards,gao2024scaling}. We use the high-scoring \citep{karvonen2024evaluating} BatchTopK SAE objective, which uses an explicit batch-level sparsity constraint \citep{bussmann2024batchtopk, leask2025sparse}
\begin{equation*}
\mathcal{L}(X)
=
\frac{1}{n}\sum_{i=1}^{n}
\left\|x_i-(W_{dec} \,\mathrm{BatchTopK}(z_i)+ b_{dec})\right\|_2^2
+
\lambda \frac{1}{n}\sum_{i=1}^{n}\|z_i\|_2^2.
\end{equation*}

In our setting, each latent dimension $z_{:,k}$ is interpreted as a candidate feature. For a fixed $(\ell, t)$, the SAE thus defines a feature set
\begin{equation}
Z^{(\ell, t)}
=
\left\{
z_{:,k}^{(\ell, t)}
\right\}_{k=1}^{m},
\end{equation}
where $z_{:,k}^{(\ell, t)} \in \mathbb{R}^{n}$ is the activation profile of feature $k$ over the dataset. This allows us to distinguish \emph{representation similarity} from \emph{feature similarity}: the former compares whole representation spaces, whereas the latter compares matched latent features extracted from those spaces.

We want to stress that not every SAE feature is human-interpretable; they can merely be a ``feature for the algorithm'' as in the work of \citet{fel2024understanding}: \emph{``Thus, in our work, a feature refers to a random scalar value extracted by a dictionary learning method on the activations.''}

\subsection{Similarity Metrics for SAE Feature Comparison}
\label{subsec:similarity_metrics}
Here, we show how we measure the similarity of the extracted features across epoch-layer pairs for our time-depth analysis. Part \textbf{C} of Figure \ref{fig:method} illustrates this feature-similarity calculation. Representation-level metrics compare the geometry of full activation spaces, but we compare individual SAE features through their activation profiles to quantify the feature-level similarity. 

For brevity, let
\begin{equation}
\mathcal{A}=(\ell, t), \qquad \mathcal{B}=(\ell', t')
\end{equation}
denote two checkpoint-layer pairs, with corresponding backbone representations $X^{\mathcal{A}}$ and $X^{\mathcal{B}}$, and SAE feature sets $Z^{\mathcal{A}}=\{z_{:,k}^{\mathcal{A}}\}_{k=1}^{m_\mathcal{A}}$ and $Z^{\mathcal{B}}=\{z_{:,j}^{\mathcal{B}}\}_{j=1}^{m_\mathcal{B}}$.

\paragraph{Transferred-feature similarity: one SAE applied across checkpoints and layers (Fixed SAE)}
The following is one part of our core methodology. Our first feature-level setting fixes a reference SAE trained at $\mathcal{A}$ and applies it to any other checkpoint-layer pair $\mathcal{B}$. Let $z_{:,k}^{\mathcal{B}\leftarrow \mathcal{A}}$ denote the activation profile of source feature $k$ from the SAE of $\mathcal{A}$ when evaluated on $X^{\mathcal{B}}$ on the dataset $\mathcal{D}$. This setting measures \emph{feature transportability}: 
%whether features discovered at $\mathcal{A}$ remain linearly readable elsewhere.
whether the same linear combination applied at a different epoch-layer pair results in a similar activation profile, i.e., instances that get a high (or low) activation at the reference epoch-layer pair, also get a high (respectively low) activation when the linear combination is applied at that other epoch-layer pair.

We compare source and transferred activations using Spearman rank correlation,
\begin{equation}
s_{\mathrm{spr}}(k;\mathcal{A}\!\to\!\mathcal{B})
=
\rho_s\!\left(z_{:,k}^{\mathcal{A}}, z_{:,k}^{\mathcal{B}\leftarrow \mathcal{A}}\right),
\end{equation}
where $\rho_s(\cdot,\cdot)$ denotes Spearman's rank correlation. 

\paragraph{Cross-SAE feature similarity: one SAE per checkpoint and layer (Independent SAE)}
Our second feature-level setting trains an independent SAE for each checkpoint-layer pair and compares the resulting features directly. For source feature $k$ in $\mathcal{A}$ and target feature $j$ in $\mathcal{B}$, we define a full cross-SAE similarity matrix using the same Spearman correlation similarity:

\begin{equation}
S_{kj,\mathrm{spr}}^{\mathcal{A},\mathcal{B}}
=
\rho_s\!\left(z_{:,k}^{\mathcal{A}}, z_{:,j}^{\mathcal{B}}\right),
\end{equation}

From this matrix, we compute row-wise best-match scores to determine the highest correlation between different epoch-layer pairs' activation profiles, possibly indicating their presence.
\begin{equation}
s^{\max}_{\mathrm{spr}}(k;\mathcal{A}\!\to\!\mathcal{B})
=
\max_j S_{kj,\mathrm{spr}}^{\mathcal{A},\mathcal{B}},
\end{equation}

\paragraph{Final similarity score as a combination of the two.}
Since both previously defined similarities have pros and cons, we reason as follows.
Spearman correlation is robust to monotone rescalings of activation magnitude. So, fixed SAE measure directly tests whether a feature learned at $\mathcal{A}$ remains stable and readable in another representation. But this score does not directly measure feature identity in the target representation. Rather, it tests whether the source dictionary remains a valid readout of the target space. A low score may therefore indicate true disappearance of the feature, but it may also arise if the same feature is present in a shifted, rotated, split, or otherwise reparameterized form. This can be mitigated with the help of our proposed independent SAE measure. Independent SAE best-match score measures whether a feature from $\mathcal{A}$ is rediscovered by the independently trained SAE at $\mathcal{B}$. This is closer to feature identity than fixed-SAE measure, since it does not assume that a single dictionary remains valid across all checkpoints and layers. Since the independent SAE score depends on both SAEs successfully learning the corresponding feature, combining it with the fixed SAE helps to achieve the desired balance between robustness to feature reparametrizations and excessive reliance on feature rediscovery. Therefore, the final method uses the maximum of the two: 
\begin{equation}
    r_k^{\mathcal{A}\to\mathcal{B}} =\max(s^{\max}_{\mathrm{spr}}(k;\mathcal{A}\!\to\!\mathcal{B}), s_{\mathrm{spr}}(k;\mathcal{A}\!\to\!\mathcal{B})) 
\end{equation}

After calculating the similarity score from $\mathcal{A}$ to each epoch-layer checkpoint, we can visualize the similarity as a heatmap and observe feature correlation across time and depth, providing insight into its evolution. Repeating this for all features $m_{\mathcal{A}}$, we can describe the evolution of features from checkpoint $\mathcal{A}$.

\section{Experiments}
\label{section:experiments}

Our depth-time view, combined with SAE-based feature evolution trajectories, allows us to investigate feature evolution with respect to where they are from, i.e., the reference epoch-layer point. We use our methodology to investigate the following questions: 

\begin{enumerate}
    \item Do features migrate from deeper to earlier layers, or from earlier to deeper layers, or in both directions?
    \item When do features stabilize across ViT layers during training?
    \item How persistent are the features learned early in the training?
    \item How layer-localized is the feature evolution during training?
    \item How class-specific are the features throughout the training?
    \item Clustering SAE features based on migration, do they form interpretable clusters?
    \item Do the answers to the previous questions depend on the dataset size?
\end{enumerate}

Now, we define the following metrics to answer the questions.

\subsection{Feature Evolution Metrics}\label{sec:migration}
Let $S_{\ell,t}^{(c)} \in \mathbb{R}$ denote the zero-capped positive similarity score of feature $c$ at layer $\ell \in \{0,\dots,L -1\}$ and epoch $t \in \{0,\dots,T\}$. By zero-capped we mean that every negative score gets replaced by zero.
%Since some statistics treat the heatmap as a mass distribution, we define the nonnegative weights
%\begin{equation} w_{\ell,t}^{(c)}=\max\bigl(S_{\ell,t}^{(c)},0\bigr). \end{equation}
For a reference point $(\ell_{\mathrm{ref}},t_{\mathrm{ref}})$, we remove the
self-similarity (being always $1$) and define a feature-specific threshold as half of the second-highest similarity score (a feature-relative threshold helps to treat features with different similarity scales accordingly).
\begin{equation}
\tau_c
=
\frac{1}{2}
\max_{(\ell,t)\neq(\ell_{\mathrm{ref}},t_{\mathrm{ref}})}
S_{\ell,t}^{(c)}.
\end{equation}
This thresholding suppresses weak background similarities and isolates the dominant similarity trajectory across time and depth to be able to capture its dynamics. We next define the above-threshold mask and filtered similarity as follows:
\begin{equation}
K_{\ell,t}^{(c)}
=
\mathbf{1}\!\left[S_{\ell,t}^{(c)}>\tau_c\right],
\qquad
\widetilde w_{\ell,t}^{(c)}
=
K_{\ell,t}^{(c)}S_{\ell,t}^{(c)}.
\end{equation}

\paragraph{Layer distribution, center, and spread.}
For each epoch, the surviving layer similarities are converted into a masked
layer distribution
\begin{equation}
P_{\ell,t}^{(c)}
=
\frac{
K_{\ell,t}^{(c)}
\exp\!\left(S_{\ell,t}^{(c)}\right)
}{
\sum_{j=0}^{L-1}
K_{j,t}^{(c)}
\exp\!\left(S_{j,t}^{(c)}\right)
},
\end{equation}
with $P_{\ell,t}^{(c)}=0$ if no layer survives. Let $r_\ell$ denote the numerical
layer coordinate (in our case $r_\ell\in \{0,\dots,11\}$). The layer center of feature $c$ at epoch $t$ and the layer spread (i.e., layer mean and standard deviation estimates)
\begin{equation}
m_t^{(c)}
=
\sum_{\ell=0}^{L-1} r_\ell P_{\ell,t}^{(c)},
\qquad
\sigma_t^{(c)}
=
\sqrt{
\sum_{\ell=0}^{L-1}
\left(r_\ell-m_t^{(c)}\right)^2
P_{\ell,t}^{(c)}
}.
\end{equation}
Low $\sigma_t^{(c)}$ indicates layer-localized similarity, whereas high
$\sigma_t^{(c)}$ indicates spread across layers.

\paragraph{Emergence and disappearance.}
The epoch-level activity signal is computed from the filtered similarity:
\begin{equation}
\alpha_t^{(c)}
=
\frac{1}{L}
\sum_{\ell=0}^{L-1}
\widetilde w_{\ell,t}^{(c)}.
\end{equation}
The emergence epoch is the first
epoch at which the feature is above threshold for $k_{\mathrm{on}}$ consecutive epochs:
\begin{equation*}
t_{\mathrm{emerge}}^{(c)}
=
\min\left\{
t:
\alpha_u^{(c)}>0
\ \text{for all}\
u\in\{t,\dots,t+k_{\mathrm{on}}-1\}
\right\},
\end{equation*}
and if $u>T$ then we define $\alpha_u^{(c)}=0$.
The disappearance epoch is the first sustained inactive run after emergence,
lasting $k_{\mathrm{off}}$ epochs, after which no later sustained
re-emergence occurs:
\begin{equation}
t_{\mathrm{disappear}}^{(c)}
=
\min\left\{
t>t_{\mathrm{emerge}}^{(c)}:
\alpha_u^{(c)}=0
\ \forall\
u\in\{t,\dots,t+k_{\mathrm{off}}-1\}
\right\},
\end{equation}
where if $u>T$ then we define $\alpha_u^{(c)}=0$.
%The active lifetime is the interval from emergence to disappearance, or to the final checkpoint if no disappearance is detected.

%\paragraph{Persistence and strength.}
%Let $\mathcal{A}^{(c)}$ denote the active above-threshold epochs during the feature lifetime. To account for non-uniform checkpoint spacing, let $\omega_t$ be the Voronoi time weight of epoch $t$. The strength/salience score is the time-weighted average activity during active epochs:
%\begin{equation} \mathrm{Strength}(c) = \frac{ \sum_{t\in\mathcal{A}^{(c)}} \omega_t a_t^{(c)} }{ \sum_{t\in\mathcal{A}^{(c)}} \omega_t }.
%\end{equation}
%The persistence fraction after emergence is the time-weighted fraction of post-emergence epochs for which $a_t^{(c)}>0$.

\paragraph{Mean layer localization.}
Layer localization is measured by the normalized entropy of the layer
distribution:
\begin{equation}\label{eq:localization}
H_t^{(c)}
=
-
\frac{1}{\log L}
\sum_{\ell=0}^{L-1}
P_{\ell,t}^{(c)}
\log\!\left(P_{\ell,t}^{(c)}+\varepsilon\right),
\qquad
\lambda_t^{(c)}
=
1-H_t^{(c)}.
\end{equation}
Thus, $\lambda_t^{(c)}\approx 1$ indicates a sharply localized layer
distribution, while $\lambda_t^{(c)}\approx 0$ indicates a spread distribution.
The mean layer localization over valid active epochs is defined as: %$\overline{\lambda}^{(c)}$.
\begin{equation}
\overline{\lambda}^{(c)}
=
\frac{
\sum_{t\in\mathcal{V}^{(c)}}
\omega_t\alpha_t^{(c)}\lambda_t^{(c)}
}{
\sum_{t\in\mathcal{V}^{(c)}}
\omega_t\alpha_t^{(c)}
}.
\end{equation}

\emph{(An example of a feature with low localization and high spread is given in Figure \ref{fig:main_fig}, part \textbf{c})}. 
%\begin{equation}
%\overline{\lambda}^{(c)} = \frac{\sum_{t\in\mathcal{V}^{(c)}} \omega_t a_t^{(c)} \lambda_t^{(c)}}{\sum_{t\in\mathcal{V}^{(c)}} \omega_t a_t^{(c)}},
%\end{equation}
%where $\mathcal{V}^{(c)}$ denotes valid active epochs, excluding epochs whose layer spread exceeds the chosen maximum-width threshold.

\paragraph{Stability.}
To measure whether a feature stays near a stable layer, we first compute the
activity- and localization-weighted layer occupancy. In our experiments we are not always using uniformly spaced epochs (for more details, look at the header of Section \ref{sec:results}), so to account for non-uniform checkpoint spacing, we assign each sampled checkpoint a midpoint-based time weight equal to the interval it represents; interior weights extend to the midpoints between adjacent checkpoints, while endpoint boundaries are extrapolated by half the adjacent checkpoint gap; let $\omega_t$ be that weight for epoch $t$
%so to account for non-uniform epoch spacing we weight each sampled epoch by the length of the interval of epochs closer to that checkpoint than to any other checkpoint;
\begin{equation}
Q_{\ell}^{(c)}
=
\frac{
\sum_{t\in\mathcal{V}^{(c)}}
\omega_t \alpha_t^{(c)}\lambda_t^{(c)} P_{\ell,t}^{(c)}
}{
\sum_{t\in\mathcal{V}^{(c)}}
\omega_t \alpha_t^{(c)}\lambda_t^{(c)}
},
\end{equation}
where $\mathcal{V}^{(c)}$ denotes valid active epochs, excluding epochs whose
layer spread exceeds the chosen maximum-width threshold $\sigma_{max}$. Intuitively, it reads as how much of the feature’s active lifetime is spent near layer $\ell$. The anchor layer is
$\ell_\star^{(c)} = \arg\max_{\ell} Q_{\ell}^{(c)}$ and for a radius $\delta$, the stability score is
\begin{equation}\label{eq:stabel}
\mathrm{Stability}_{\delta}(c)
=
\sum_{\ell:\, |r_\ell-r_{\ell_\star^{(c)}}|\leq \delta}
Q_{\ell}^{(c)}.
\end{equation}
High stability indicates that the feature remains close to one layer during its active lifetime. \emph{(An example of a stable feature with high mean layer localization is in Figure \ref{fig:main_fig}, part \textbf{b}).}

\paragraph{Signed active drift.}
Let $\mathcal{V}^{(c)}_{\mathrm{start}}$ and
$\mathcal{V}^{(c)}_{\mathrm{end}}$ denote the first and last fractions of valid
active epochs. The start and end layer distributions are $Q_{\mathrm{start}}^{(c)}$
%\begin{equation} Q_{\mathrm{start}}^{(c)} =\frac{ \sum_{t\in\mathcal{V}^{(c)}_{\mathrm{start}}} \omega_t a_t^{(c)}\lambda_t^{(c)}  _{\ell,t}^{(c)} }{\sum_{t\in\mathcal{V}^{(c)}_{\mathrm{start}}}\omega_t _t^{(c)}\lambda_t^{(c)}}, 
%\end{equation}
and $Q_{\mathrm{end}}^{(c)}$. The signed mean active drift is
\begin{equation}\label{eq:drift}
D_{\mathrm{mean}}^{(c)}
=
\sum_{\ell=0}^{L-1} r_\ell Q_{\mathrm{end},\ell}^{(c)}
-
\sum_{\ell=0}^{L-1} r_\ell Q_{\mathrm{start},\ell}^{(c)}.
\end{equation}
Positive values indicate movement toward deeper layers, while negative values
indicate movement toward earlier layers. \emph{(An example of feature migration towards earlier layer is in Figure \ref{fig:main_fig}, part \textbf{a})}. 

\paragraph{Feature classes.}\label{par:mig_thres}
In our experiments, a feature is classified as \emph{spread} if mean layer localization $\overline{\lambda}^{(c)}$ is smaller than $0.5$. We classify a feature as migratory only when it's not spread and its estimated displacement exceeds one complete layer, meaning a feature is classified as \emph{deeper-migratory} if $D_{\mathrm{mean}}^{(c)} > 1$ and \emph{earlier-migratory} if $D_{\mathrm{mean}}^{(c)} < -1$. A feature is classified as \emph{stable} if it is not migratory nor spread, $|D_{\mathrm{mean}}^{(c)} |< 0.5$ and $\mathrm{Stability}_{1}(c)\geq 0.7$. The former condition requires the net displacement of a stable feature to remain below half a layer. This is combined with $\mathrm{Stability}_{1}(c)\geq 0.7$, requiring at least 70\% of the activity- and localization-weighted layer occupancy to lie within one layer of the feature's anchor layer. Thus, the stability criterion captures both low net displacement and sustained concentration around a common layer.  We deliberately leave a margin between the stable and migratory regimes: shifts below half a layer are considered negligible, whereas migration requires a displacement exceeding one full layer. Intermediate displacements are not assigned to either category. These cutoffs are not intended as theoretically unique decision boundaries but provide conservative and interpretable operational categories in units of layers and normalized occupancy.

%In addition, we calculated the mean label entropy \citep{lim2024sparse} from the feature label statistics as
%\begin{equation*}
%\overline{H}_{\mathrm{label}} =\frac{1}{|\mathcal{C}|}\sum_{c\in\mathcal{C}}H_{\mathrm{label}}^{(c)}.
%\end{equation*}

\begin{figure*}
    \centering
    \includegraphics[width=\textwidth,trim={0 474 0  0},clip]{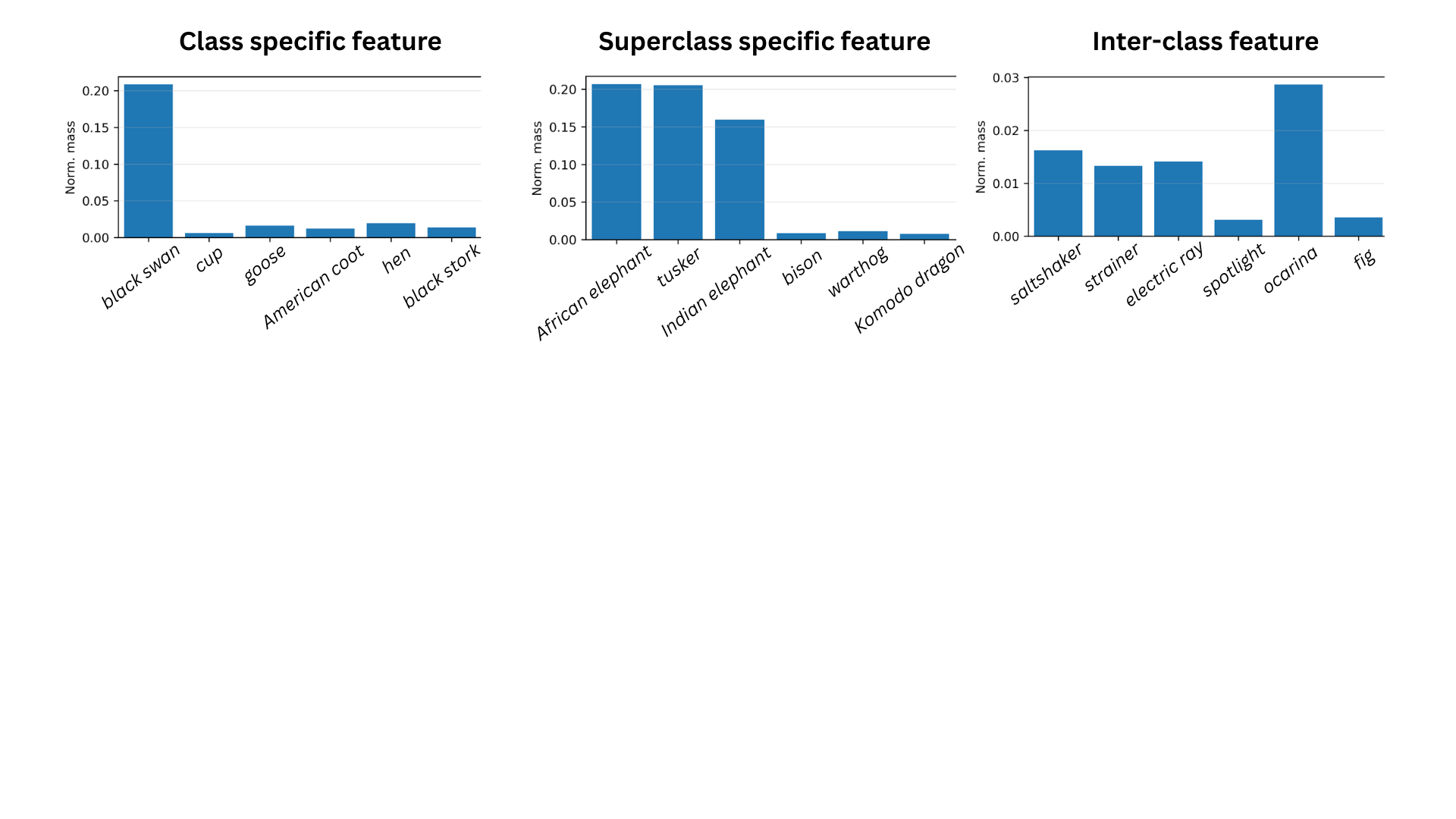} %change this to ..types_3 for high res
    \caption{Feature-types based on ImageNet classes. Human-understandable SAE features are observed under 3 main categories: class-specific, superclass-specific and inter-class (see maximally activating images \citep{zeiler2014visualizing} for each feature in Appendix).}
    \label{fig:feature_types}
\end{figure*}

\paragraph{Feature validation}\label{par:feat_val} To answer the question of whether SAE features cluster based on their migratory behavior, we need to somehow label the features. We follow \citep{pach2025sparse} and interpret a set of SAE features manually. Our manual effort revealed that SAE features can usually be put under one of the categories: \emph{class-specific}, \emph{superclass-specific}, \emph{inter-class}, \emph{not human-understandable} (at least not immediately). We automate the process partly as follows. We classify each feature by examining the normalized ImageNet class distribution of the images in the test set where that feature is active. If one class is clearly dominant, meaning that the normalized activation for the second-highest class is less than half of that of the highest class, the feature is labeled a class-specific feature (see the left of Figure \ref{fig:feature_types}). If there is no single dominant class, we check whether a small group of 2 to 10 top-ranked classes has relatively similar normalized activations and is followed by a large drop, defined as the next class having less than half the normalized activation of the lowest-ranked class in that group. This indicates that the feature may be shared by a small semantic group rather than a single class. To confirm this, we utilise the semantic hierarchy of WordNet \citep{fellbaum1998wordnet}: the top classes must share an immediate hypernym. If they do, the feature is labeled a superclass feature (see the middle of Figure \ref{fig:feature_types}). If neither condition is satisfied, meaning a feature is not clearly associated with one class or with a tightly related group of classes, we return to our manual search and check if any of the remaining features indicate a clear pattern, i.e., whether it is an inter-class high-level feature (see the right of Figure \ref{fig:feature_types}).

\iffalse
\subsection{Probing features across layers and training epochs}
In this view of the ViT traning we probe CLS token in different ViT layers at different training epochs. This gives us a layer wise longitudinal view of how well these featrues are linearly represented in the network at different stages of training. \todo{add probing method} To get a per-feature overview, we first visualize the test set probing accuracy per feature at each layer and training epoch. Further we quantify the drift in features across layers as the training progresses measured by the change in center of mass for the accuracy across layers at each epoch. Plots comparing the center of mass 
\fi

\subsection{Setup}

\paragraph{ViT training}

We use two ImageNet-based datasets: ImageNet-1k \citep{imnet1k} and its balanced superclass dataset \emph{Mixed 10} adapted from \citet{robustness}.

We use the vanilla architecture of the ViT \citep{dosovitskiy2020vit} to assess its feature evolution, as it is a primary building block of modern vision models, e.g., DINOv2 and CLIP. More precisely, we exemplify our framework on ViT-Tiny \citep{touvron2021training}, but use 12 heads as suggested by \citep{wang2023closer}. \citet{wang2023closer} also showed that the representations of ViT-Tiny and ViT-Base models trained from scratch look similar using CKA, allowing us to generalize across ViTs with our analysis on the computationally efficient ViT-Tiny.

We trained a ViT-Tiny on ImageNet-1k following the setup from \citep{wang2023closer}, using \(12\) transformer blocks, patch size \(16\), embedding dimension \(192\), and \(12\) attention heads. Training was performed for \(300\) epochs with AdamW (\(\beta_1=0.9\), \(\beta_2=0.999\), weight decay \(0.05\)) and a cosine learning-rate schedule \citep{loshchilov2016sgdr} with \(5\) warmup epochs. The base learning rate was set to \(10^{-3}\) and scaled linearly  with the global batch size  as \(\mathrm{lr}=10^{-3}\cdot \tfrac{B}{256}\) \citep{bao2021beit}. We used batch size \(2048\) and mixup \citep{zhang2017mixup} with \(\alpha=0.2\). Input images were resized to \(224\times 224\), and training augmentations followed the paper-inspired recipe with random resized crops, horizontal flipping, color jitter (\(0.3\)), and RandAugment \citep{cubuk2020randaugment}. We additionally used layer-wise learning-rate decay with coefficient \(0.75\), mixed-precision training, and standard ImageNet normalization. To speed up the dataloader, we used FFCV \citep{leclerc2022ffcv}. For the ImageNet-1k superclass dataset \emph{Mixed 10}, we used the same training recipe but with a smaller batch size \(512\). The achieved accuracies were for ImageNet-1k $72.55\%$ and for \emph{Mixed 10} $86.37\%$.

\paragraph{SAE training}
Recent work suggests that the distinction between CLS and patch tokens is not clean: standard ViTs may already treat them differently through normalization and other implicit mechanisms, despite nominally shared computation \citep{marouani2026revisiting}. \citet{darcet2023vision} also suggest that global information can leak into patch tokens, complicating the interpretation of patch-stream features as purely local. For clarity of interpretation, we limit ourselves in this paper to the CLS tokens.

Therefore, we train SAEs on top of CLS tokens. \citet{gao2024scaling} suggests training SAE until convergence, but in our setup, such computation was not feasible, and we fixed the number of epochs to $150$ (at this point, the loss started to plateau during hyperparameter search). For ImageNet-1k, we limited the training set size via random sample to be $N_{train}=100\,000$ uniformly distributed among $1000$ classes, and for testing we used the original ImageNet-1k validation set as a test split ($N_{test}=50\,000$). For \textit{Mixed 10} dataset, we used the entire training split ($N_{train}=77\,060$) and test split ($N_{test}=2700$). We used the same SAE settings for all the checkpoints and both datasets: batch size 4096, sparsity expansion coefficient 5, BatchTopK with $K=32$, L2 regularization 0.05.

\iffalse
\paragraph{Probing setup}
 For probing, we use the CUB birds dataset with attributes \citep{wah_branson_welinder_perona_belongie_2023}.
 We trained separate binary logistic-regression probes for each attribute $c \in \mathcal{C}$ using the corresponding binary labels $y^c_i$. To test
generalization beyond class-specific memorization, probe training and evaluation were performed using class-held-out splits: bird species appearing in the probe
test set were excluded from the probe training set. Because CUB attributes are often highly imbalanced, performance was measured using balanced accuracy rather
than raw accuracy. Positive/negative filtering... L2 regularization, class weighting... Solver, convergence tolerance, max iterations, repeated random splits/seeds...
\fi

\subsection{Results}\label{sec:results}
We report the results for both datasets, if they are substantially different, otherwise, we show only the ImageNet-1k results, since its larger test set is less noisy. The results for the migration metrics defined in Section \ref{sec:migration} are presented with a high-frequency window of $[\text{reference epoch} - 50; \text{reference epoch} + 50]$ around the reference epoch and low-frequency elsewhere (i.e., every 10th epoch). For the reference epochs smaller than $50$ or greater than $250$, the window was set to $[0,100]$ and $[200,300]$, respectively. Reference epochs were chosen mainly from the beginning of the training, as it is when the model is learning more rapidly, and more sparsely after that, resulting in reference epochs $[0,100]\cup\{110,120,\dots,280\}\cup[281,300)$.

\begin{figure}[ht]
  \centering
  \includegraphics[width=\linewidth]{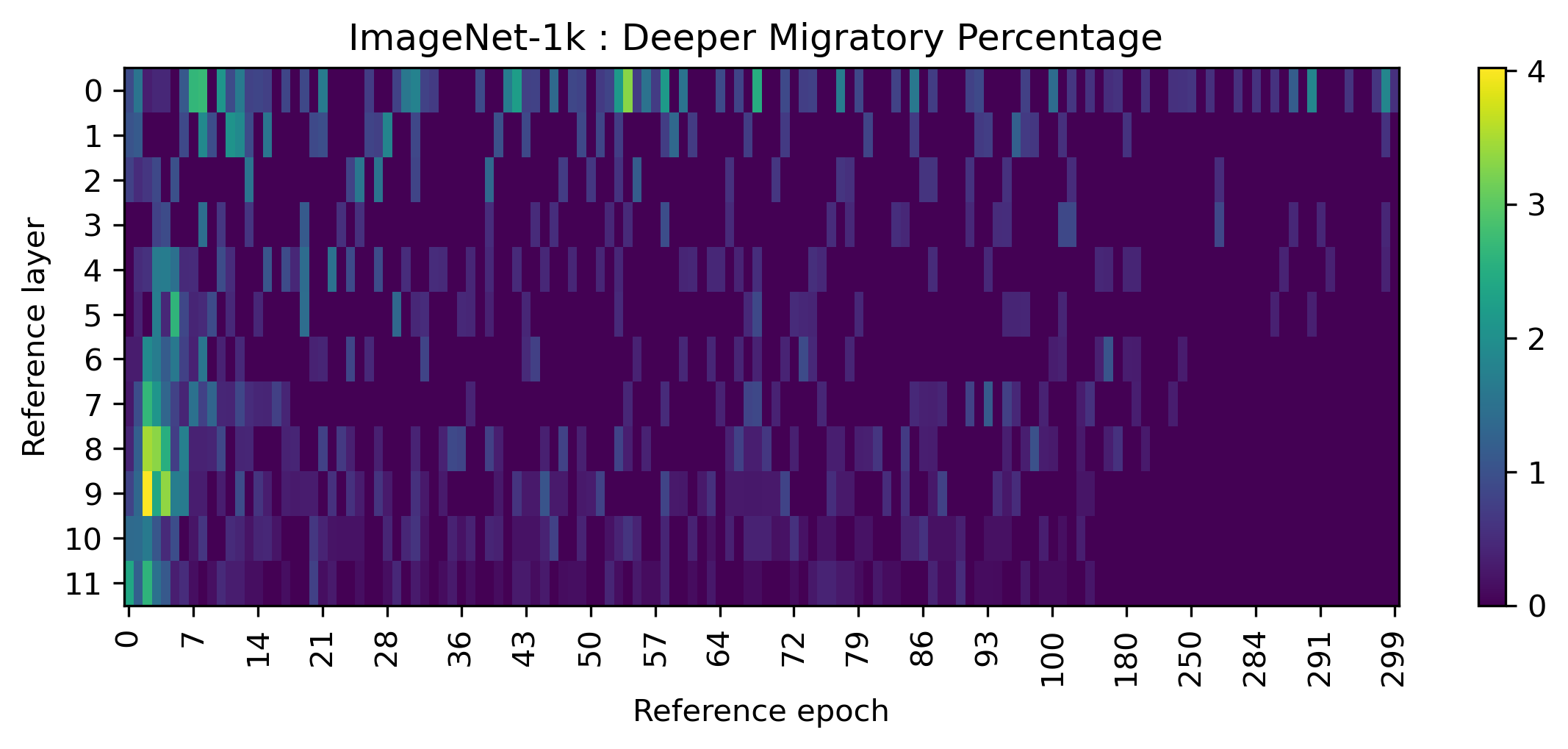}
  \caption{Percentage of features migrating towards deeper layers.}\label{fig:deeper} 
\end{figure}

\paragraph{Do features migrate from deeper to earlier layers, or from earlier to deeper layers, or in both directions?} We conducted several experiments to answer this question. Firstly, we used Eq. \ref{eq:drift} with the given thresholds from Section \ref{sec:migration} and calculated the percentage of deeper migrating features among all the reference points; the resulting heatmap is depicted in Figure \ref{fig:deeper}. We can see that deeper migration is relatively rare, peaking at $4\%$. The few visible peaks occur mostly in early checkpoints, suggesting that movement toward deeper layers is not a dominant mode of feature reorganization.

\begin{figure}[ht]
    \centering

    \begin{subfigure}{0.5\textwidth}
        \centering
        \includegraphics[width=\linewidth]{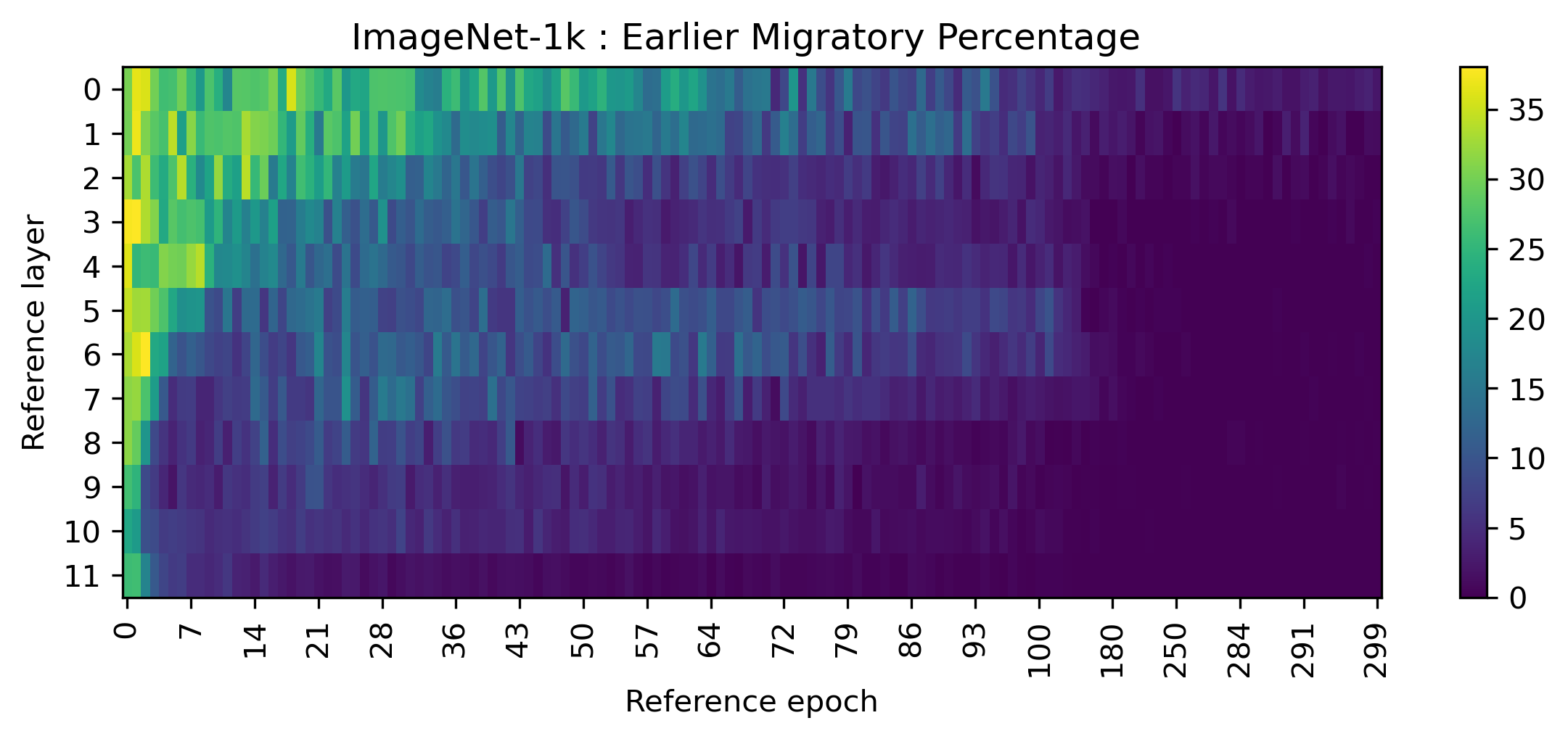}
        \label{fig:earlier_s8}
    \end{subfigure}
    \hfill
    \begin{subfigure}{0.5\textwidth}
        \centering
        \includegraphics[width=\linewidth]{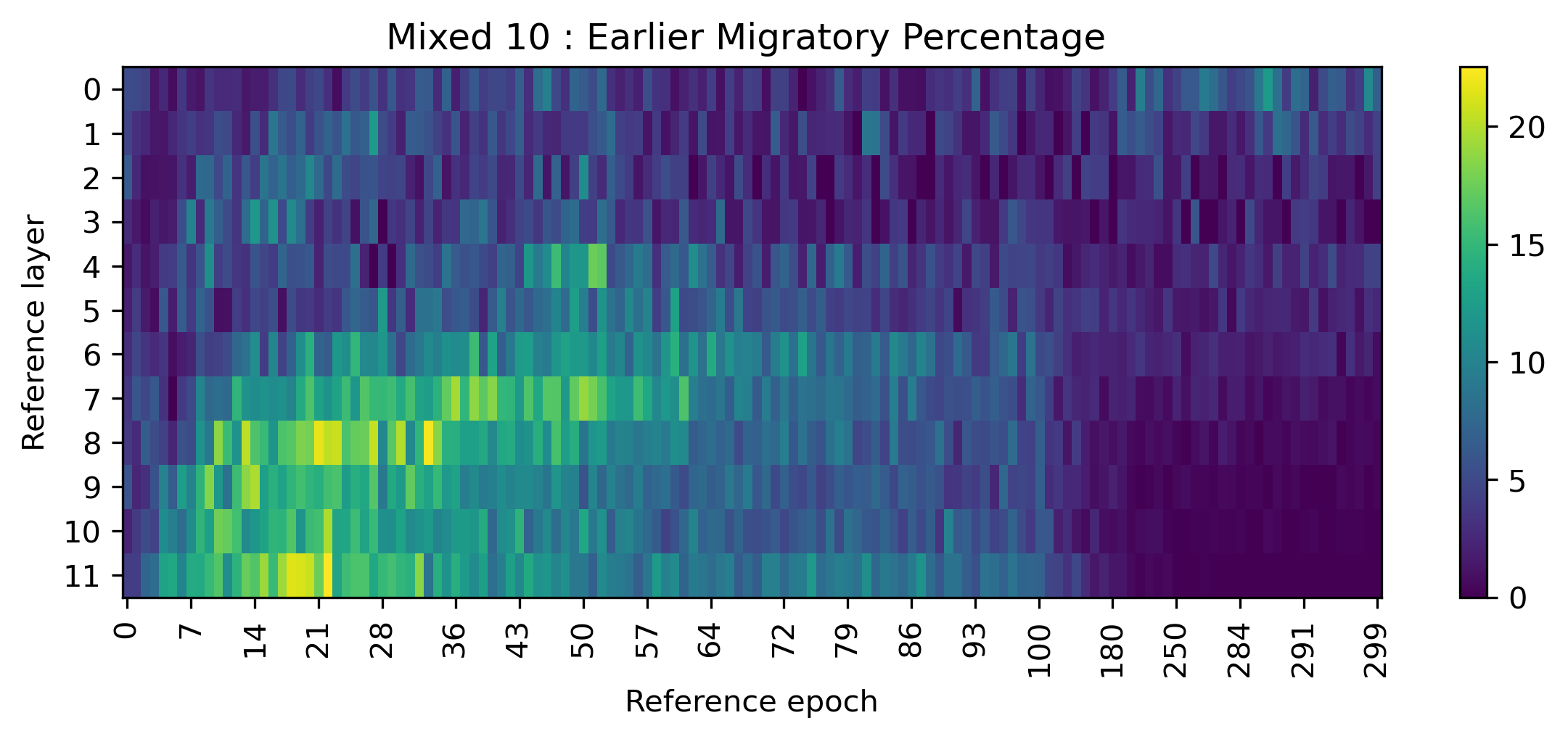}
        \label{fig:earlier_s7}
    \end{subfigure}

    \caption{Percentage of features migrating towards earlier layers.}
    \label{fig:earlier}
\end{figure}

Analogously, the resulting heatmap for the percentage of features migrating towards earlier layers is given in Figure \ref{fig:earlier}, which shows that, for both datasets, it is most prevalent primarily in the first part of the training. The main difference is in the layers affected: early layers in the case of ImageNet-1k and later layers in the case of Mixed 10. For clarity, features from layer 0 can still migrate towards earlier layers, as the definition \ref{eq:drift} does not require that feature drift necessarily begin at the reference layer, but rather where it is most active, which can, in fact, be later layers.

\begin{figure}[ht]
  \centering
  \includegraphics[width=\linewidth]{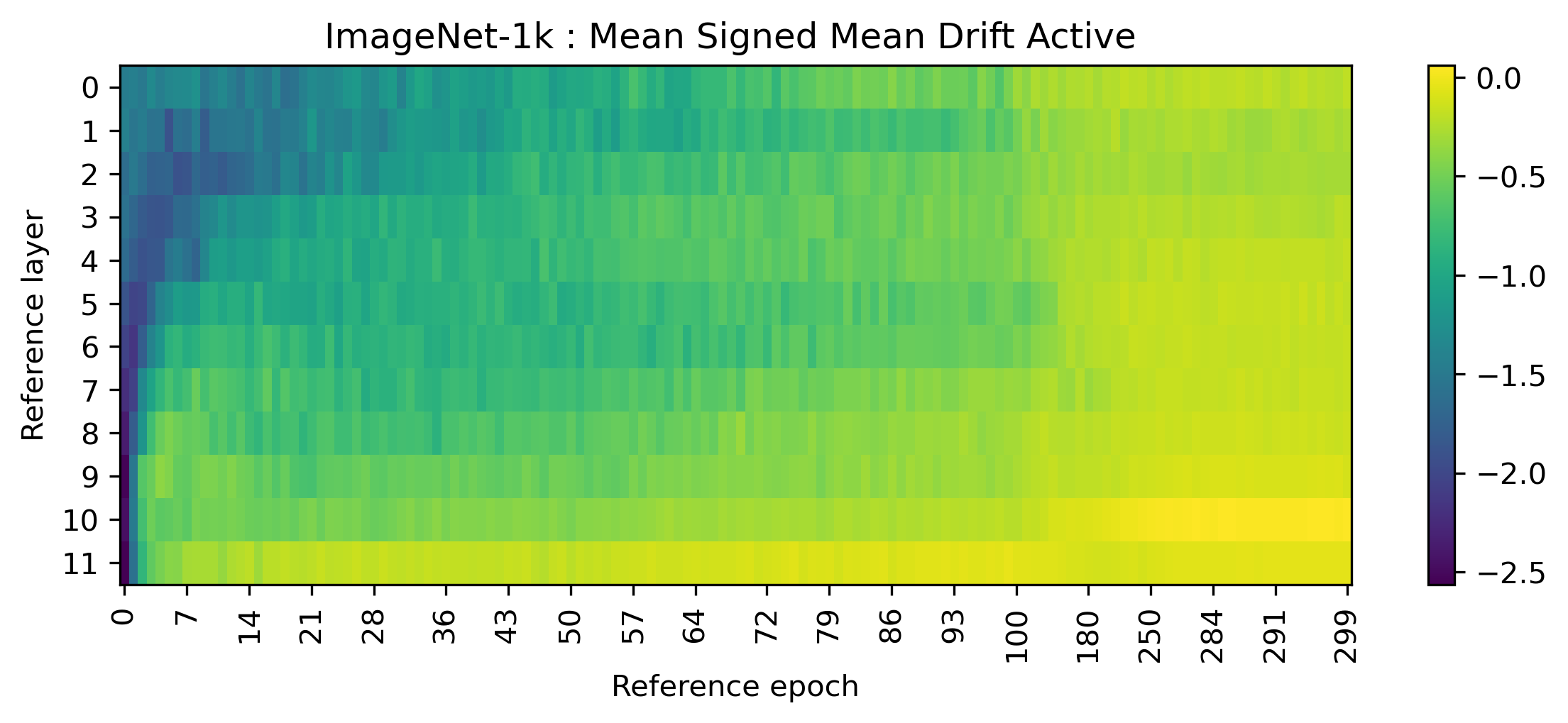}
  \caption{Mean signed active drift. For each reference epoch--layer pair, feature-level quantities are averaged across features. Average signed displacement of the layer center, in layer-index units; negative values indicate movement toward earlier layers.}\label{fig:drift_magnitude}
\end{figure}

Neither of the previous plots says much about the gravity of migration; e.g. whether migration is across 1 or even 5 layers. We depict the average of signed mean drift in Figure \ref{fig:drift_magnitude}. We see that deeper layers and the later training phase are, on average, less affected by migration. On the contrary, earlier layers and training show higher migration towards earlier layers, reaching approximately $-2.5$ layers in the regions with strongest migration.

\begin{figure*}[ht]

  \centering
  \includegraphics[width=\linewidth]{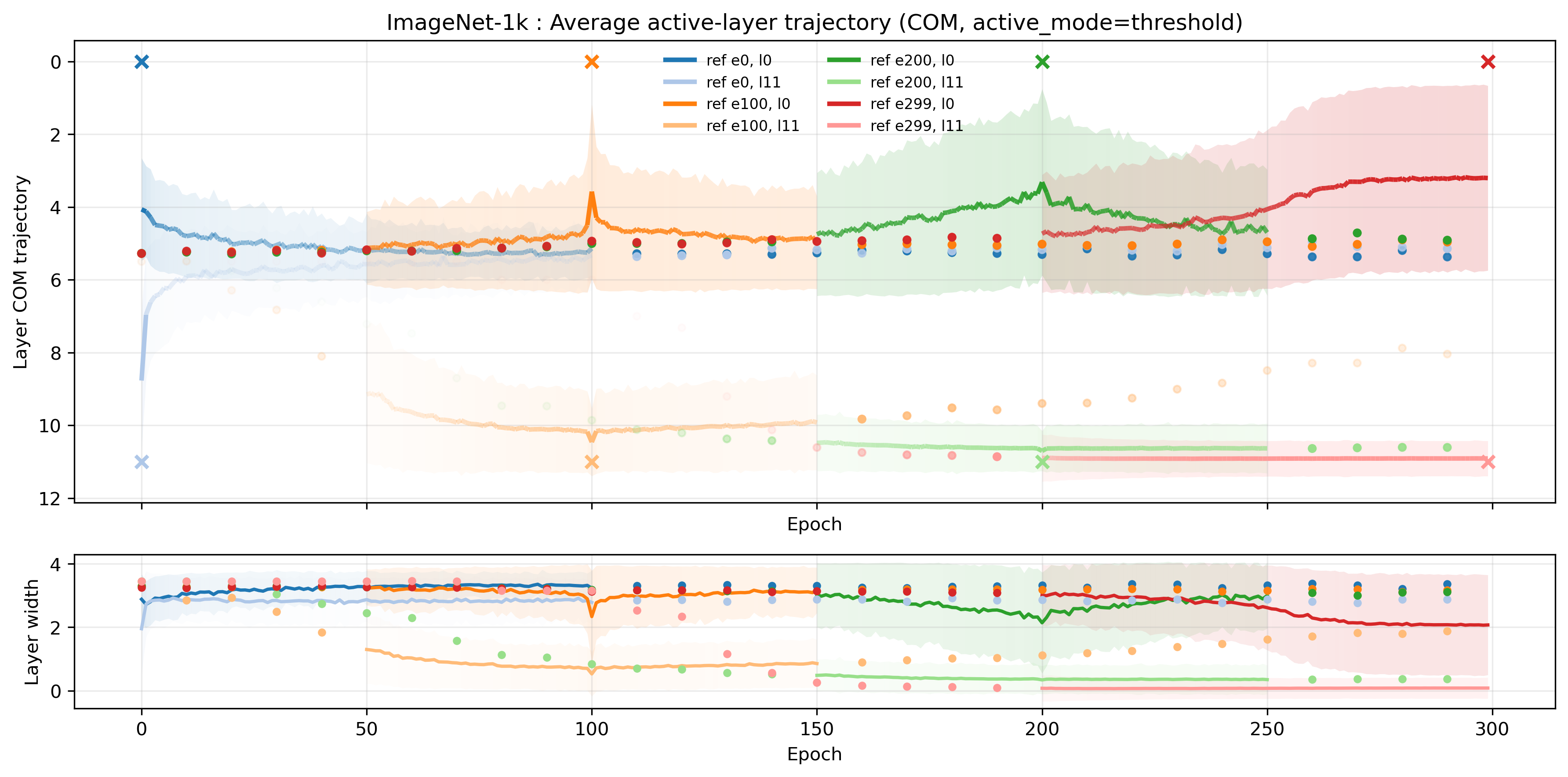}
  \caption{Layer center-of-mass (COM) $\pm$ standard deviation trajectories averaged over all the features of selected reference checkpoints (0,100,200 and 300). The high-frequency epochs are depicted as a solid line around the reference checkpoints (marked as x) and the low-frequency epochs as dots. On the figure below, the average layer width $\pm$ standard deviation is depicted. The averaging is carried out only on active epochs per feature and therefore color fading signifies the lack of active features in that area.}\label{fig:traj}
\end{figure*}

In Figure \ref{fig:traj}, we depict the layer center-of-mass (COM) trajectory averaged over all the features per reference point. In the figure, notably, after the first epoch, we can see migration predominantly towards deeper layers from layer 0 (but rarely exceeding 1 layer) and towards earlier layers from layer 11. The features from the last layer in epoch 300 are stable and have little deviation around the COM, whereas the first layer is less stable and can experience migration towards earlier layers. Epochs 100 and 200 show a peak at the reference location -- this may be due to the localization of similarity at the reference layer and diffusion or migration towards earlier layers thereafter. The peak is smoother for the first layer at epoch 200, indicating feature stabilization. The last layer at the same epochs shows more stability, and COM captures the reference layer better. Curiously, the average layer width for features referenced from the first layer increases over training, indicating that their similarity mass becomes less layer-localized.

\begin{figure}[ht]
  \centering
  \includegraphics[width=\linewidth]{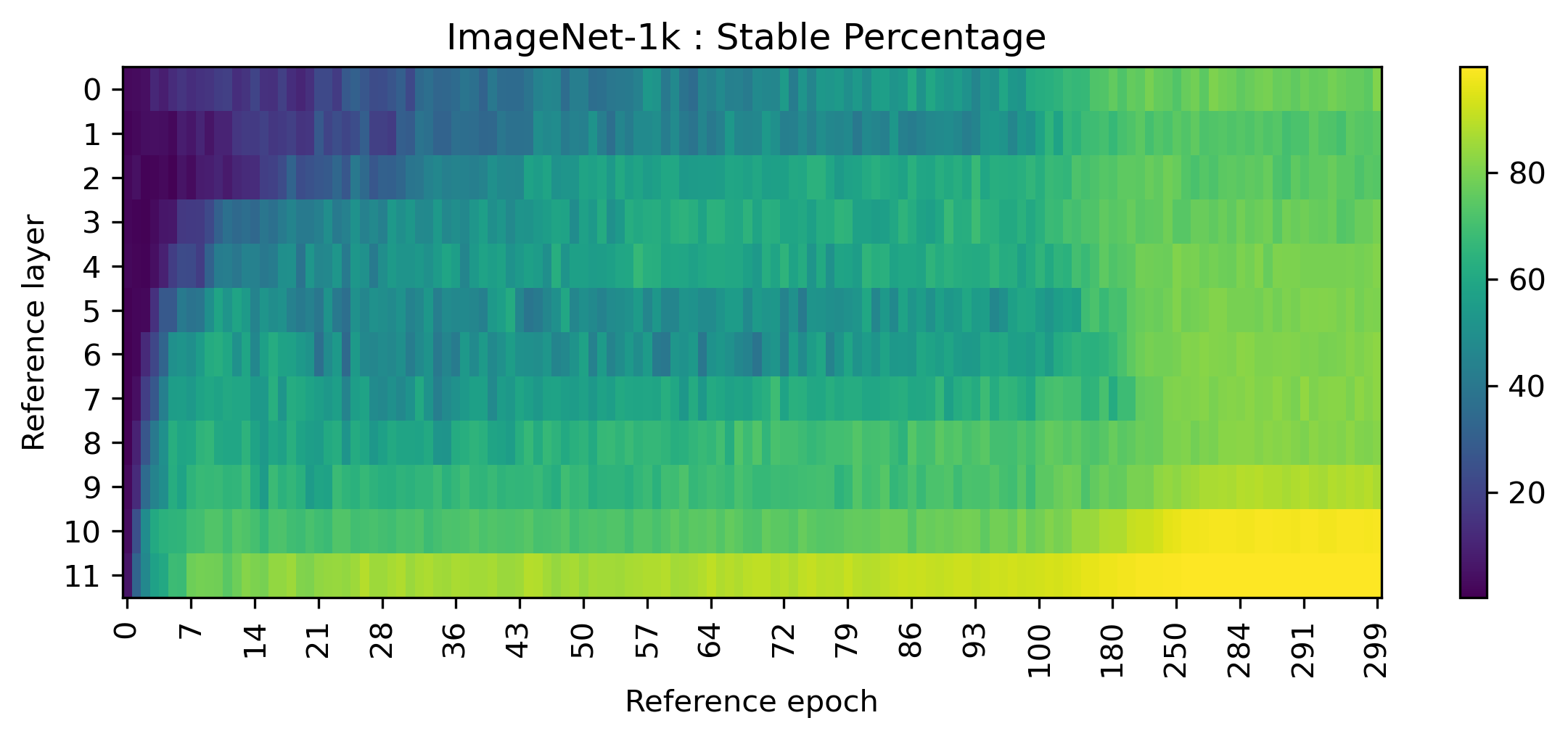}
  \caption{Percentage of stable features. }\label{fig:stability}
\end{figure}

\begin{figure}[ht]
  \centering
  \includegraphics[width=\linewidth]{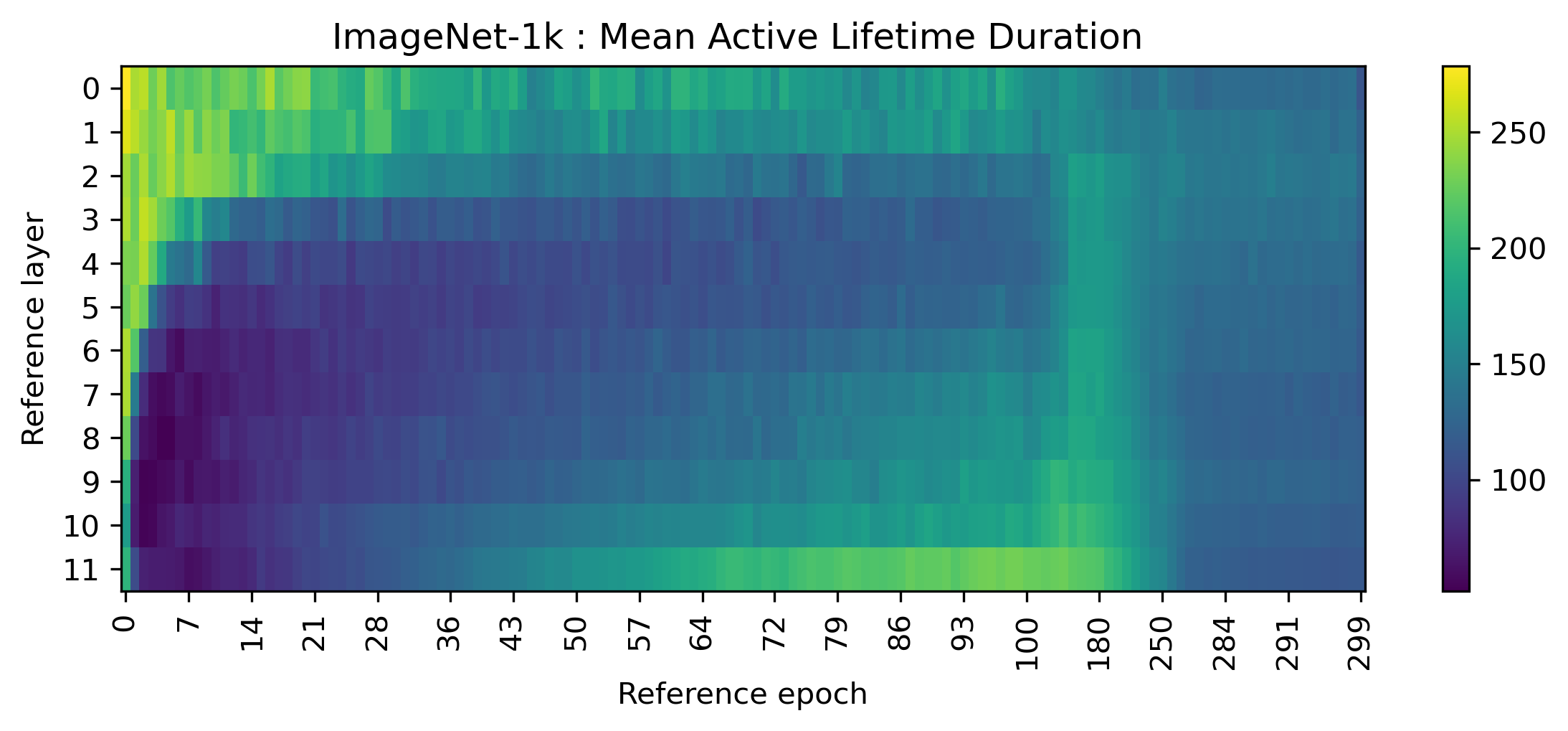}
  \caption{Mean active lifetime of features, given in number of epochs.}\label{fig:disap}
\end{figure}

\paragraph{When do features stabilize across ViT layers during training?}
This question is partially answered in the previous paragraph. We used Equation \ref{eq:stabel} with the given thresholds from Section \ref{sec:migration} and counted the percentage of stable features from all the reference points; the resulting heatmap is depicted in Figure \ref{fig:stability}. Naturally, it correlates visibly with Figure \ref{fig:drift_magnitude} of average migration magnitude but in this case we can also see how many of the features are considered stable. Firstly, the deeper layers stabilize, and throughout training, earlier layers stabilize as well, but not to the same extent as the deeper layers.

\paragraph{How persistent are features learned early in the training?}
In Figure \ref{fig:disap}, the mean active lifetime of features in epochs (the mean number of epochs between emergence and disappearance) is depicted per epoch-layer pair. Interestingly, some earlier layers and epochs exhibit long persistence throughout the training, even though they are not stable, as shown in Figure \ref{fig:stability}. Surprisingly, deeper and intermediate layers in early training tend to have a short lifetime. The average active lifetime for late-training-phase features seems stable.

\paragraph{How layer-localized is the feature evolution during training?}

\begin{figure}[ht]
  \centering
  \includegraphics[width=\linewidth]{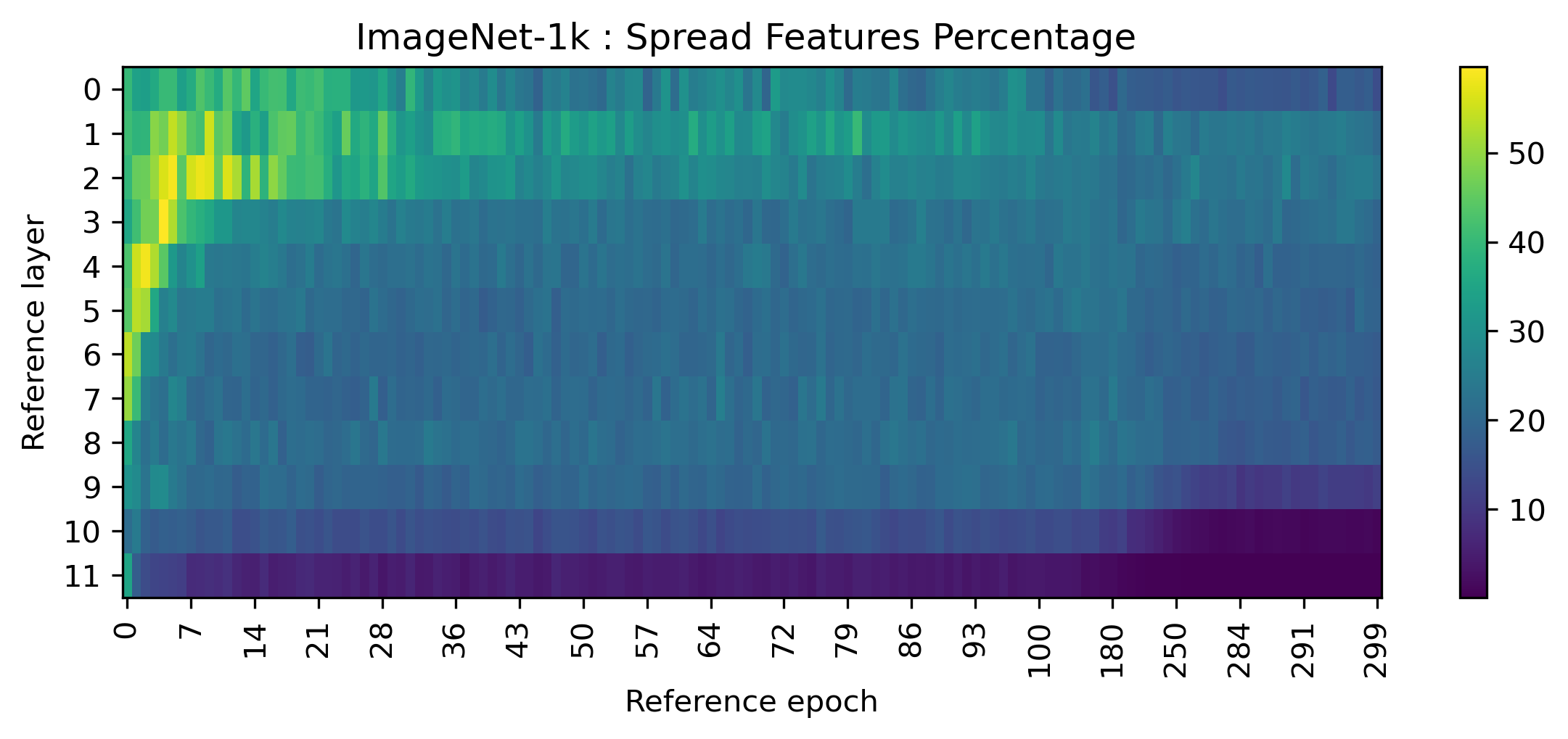}
  \caption{Percentage of features that are widely spread across layers.}\label{fig:spread}
\end{figure}

To measure the layer localization, we used Equation \ref{eq:localization} with the given thresholds from Section \ref{sec:migration} and counted the percentage of features that are widely spread across layers from all the reference points; the resulting heatmap is depicted in Figure \ref{fig:spread}. Interestingly, widely spread features occur throughout training, mainly at the beginning and in the earlier layers. Therefore, combined with the percentage of stable features, there is no more significant migration left at the end of training.

\begin{figure}[ht]
  \centering
  \includegraphics[width=\linewidth]{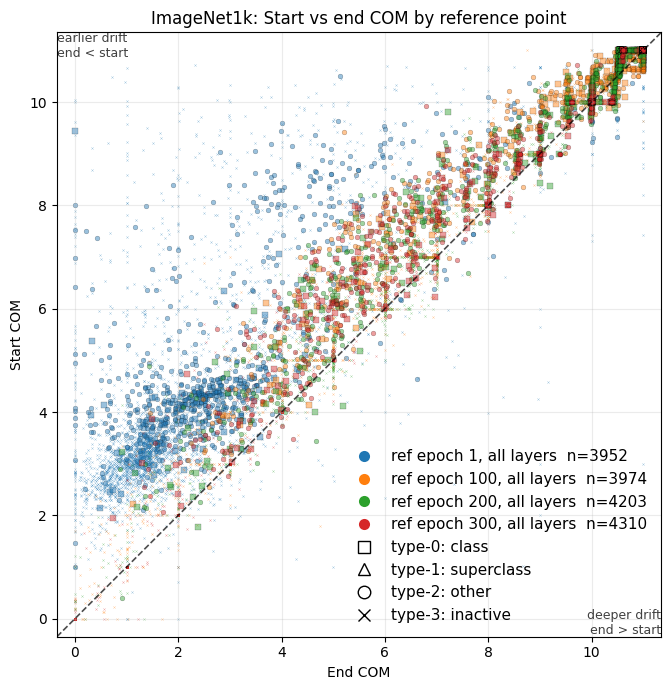}
  \caption{Feature migration analysis by layer center-of-mass (COM). Most of the features migrate towards earlier layers.}
  \label{fig:scatter}
\end{figure}

\paragraph{Clustering SAE features based on migration: do they form interpretable clusters?} To further investigate the feature migration patterns, we provide a layer center-of-mass (COM) analysis in Figure \ref{fig:scatter}, which shows the change in COM for features from different reference points. As can be seen from the figure, a vast majority of the features are migrating towards earlier layers, suggesting that they start emerging in later layers in early training, then move towards earlier layers, e.g., the behavior we see in Figure \ref{fig:main_fig} part \textbf{(a)}. Moreover, layer COMs for features referenced to very early training checkpoints, i.e., epoch 1, mostly cluster together and move roughly from layers 3-5 to layers 1-3. Other reference features from later training, such as at epochs 100, 200 or 300, are mostly active after the 4th layer in the early training and move 1-3 layers up later towards the end of the training. The presented features are also labeled automatically as mentioned in Section \ref{sec:migration}; however, we do not observe any notable correlation between their types, e.g., class-specific, superclass-specific, or others, and their migration behavior. 

\begin{table}[t]
\centering
\begin{tabular}{ccccc}
\hline
\textbf{Reference epoch} & \textit{class} & \textit{superclass} & \textit{other}  & \textit{total (active)} \\
\hline
1   & 70 & 0 & 755 & 825  \\
100 & 259 & 22  & 678 & 959  \\
200 & 401 & 35 & 872 & 1308 \\
300 & 461 & 42 & 903 & 1406 \\
\hline
\end{tabular}
\caption{Feature-type counts by reference epochs presented in Figure \ref{fig:scatter}.}
\label{tab:concept-type-counts}
\end{table}

\begin{figure}[ht]
    \centering

    \begin{subfigure}{0.48\textwidth}
        \centering
        \includegraphics[width=\linewidth]{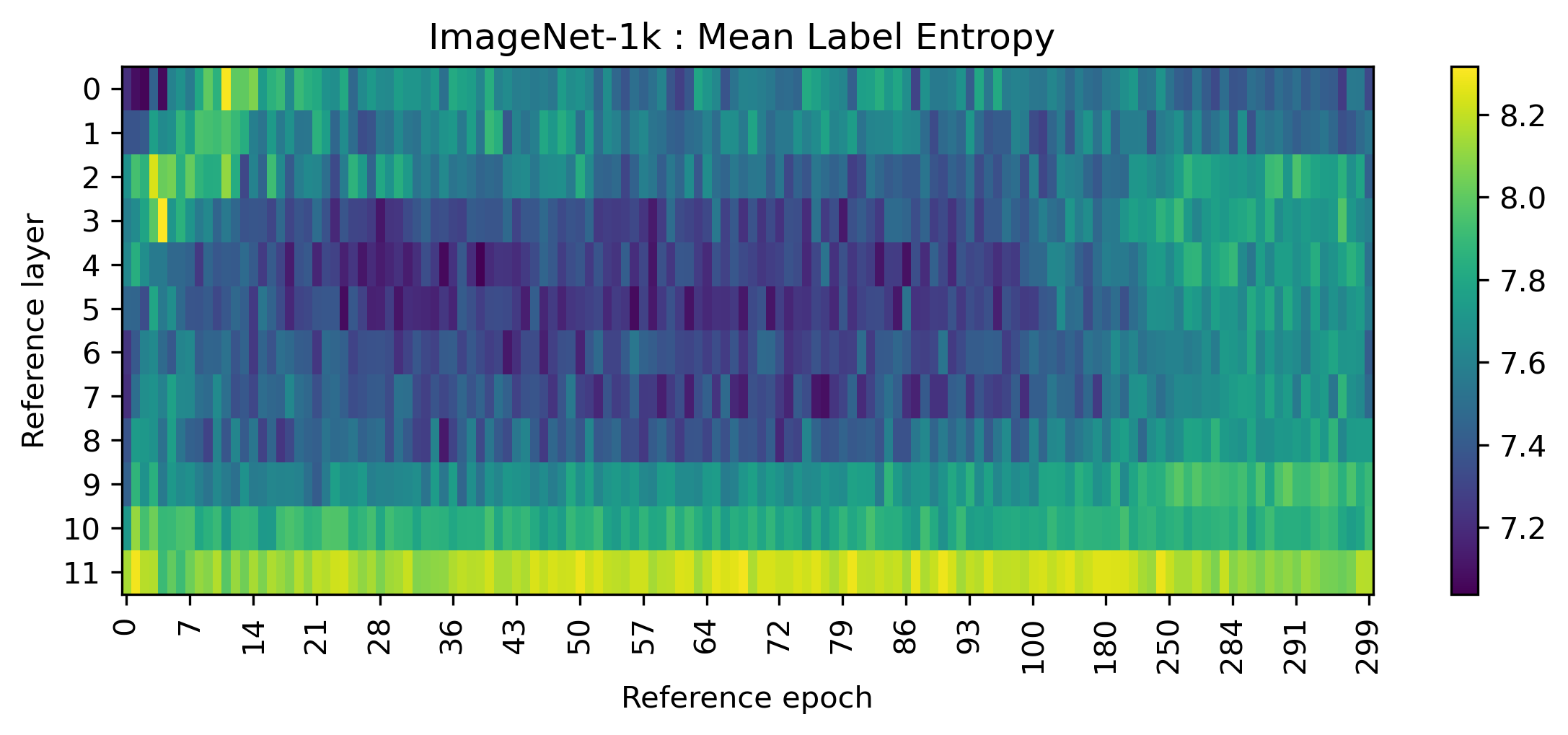}
        \label{fig:label_entopys8}
    \end{subfigure}
    \hfill
    \begin{subfigure}{0.48\textwidth}
        \centering
        \includegraphics[width=\linewidth]{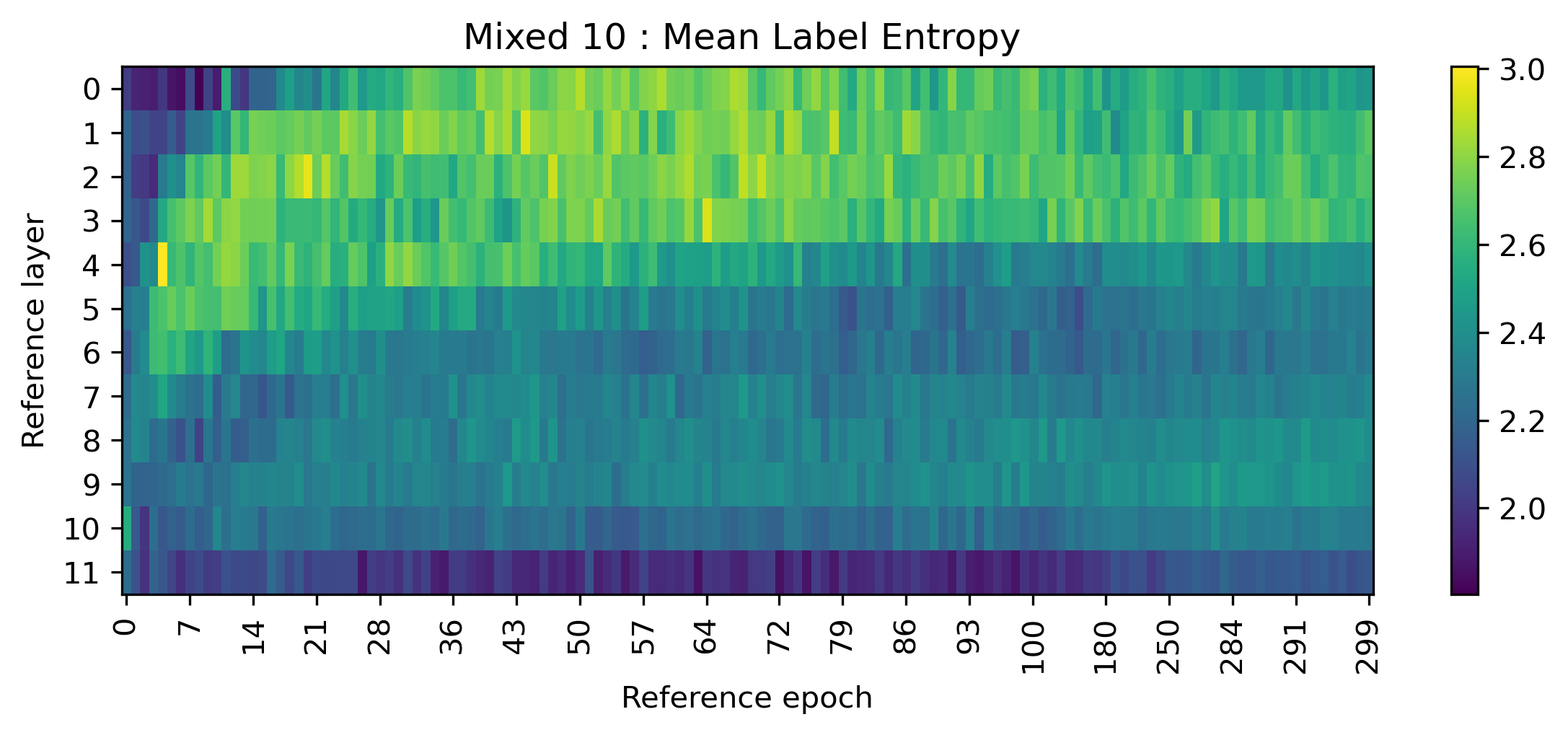}
        \label{fig:label_entropys7}
    \end{subfigure}

    \caption{The difference between two datasets in label entropy. Low label entropy indicates label-selective features, whereas high label entropy indicates features that activate across many labels.
}
    \label{fig:label_entropy}
\end{figure}

\paragraph{How class-specific are the features throughout the training?}
Furthermore, we measured the label entropy \citep{lim2024sparse} of our features based on the image labels for which the feature is activated, resulting in Figure \ref{fig:label_entropy}. We find that there is a stark difference between the datasets, ImageNet-1k having high label entropy in higher layers whereas \emph{Mixed 10} having high entropy, on the contrary, in the lower layers consistently.

Additional results and experiment details are presented in the Appendix~\ref{appendix:all}.

\section{Discussion and Conclusion}

Our observations of long feature lifetime in the early layers and prevalent migration towards earlier layers are consistent with prior claims by \citet{yun2021transformer} and \citet{fel2024understanding}. More precisely, our finding that early layer features have a long active lifetime relates to \citet{yun2021transformer}, who claimed for language Transformers (BERT \citep{devlin2019bert}) that, due to residual connection, we should expect ``\textit{if a vital feature is learned in the beginning layers, it won’t disappear in later stages. Instead, it will be carried all the way to the end with gradually decayed weight since many more features would join along the way.}'' Furthermore, in Figure \ref{fig:traj}, the average layer width increased over training, which might be due to the residual connections gradually having more effect on deeper layers. A similar observation about the role of residual connections is made by \citet{raghu2021vision}, who notice that ViTs exhibit more uniform representations across depth, make use of global information earlier, and are more strongly shaped by residual connections than CNNs. In Figure \ref{fig:scatter}, we did not observe any clustering of feature types by migration, but we do see a progression across feature types over training time: the later the epoch, the more class- and superclass-specific features there are (see Table \ref{tab:concept-type-counts}). This relates to the finding by \citep{bau2017network} in CNNs that low-level features are learned earlier in training. \citet{sharon2024does} analyzed how representations change through training in CNNs and ViTs, showing that layers can evolve differently across regimes and that deeper layers may change more substantially during prolonged fitting -- as prolonged fitting was not part of our routine, we can neither confirm nor refute their finding, but on the contrary, we did see more instability and migration in earlier layers rather than deeper ones. 

Our finding about the label entropy in the previous section, depicted in Figure \ref{fig:label_entropy} can possibly be explained by the expansion coefficient of the SAE. Since the image representations have dimensionality $192$ and are mapped into a sparse layer via SAE of dimensionality $5\times192 = 960$, then SAE cannot afford to recover only single-class-specific features for ImageNet-1k. The opposite seems to be the case for the \emph{Mixed 10} dataset. Therefore, this might have implications for the selection of the SAE expansion coefficient when class-specific features are desirable.
%In addition, we have not seen anyone demonstrate that SAEs can recover human-readable features from plain ViT without any special training recipe, so we also manually confirmed this possibility.

Our work can be described as feature-level readouts of weight-space training dynamics. A substantial line of work studies neural network training as a trajectory through parameter space, using tools such as interpolation from initialization to solution \citep{goodfellow2014qualitatively}, loss-landscape visualization \citep{li2018visualizing}, mode connectivity \citep{garipov2018loss}, stochastic weight averaging \citep{izmailov2018averaging}, and curvature-based analyses of optimization dynamics \citep{cohen2021gradient}. Our method instead studies an interpretable feature-level projection of the same training trajectory and, therefore, is complementary to this weight-space perspective. Rather than measuring distances, barriers, curvature, or connectivity between checkpoints in parameter space, we ask how the model’s movement through parameter space is reflected in the development of interpretable internal features. One can view our method as defining a feature-level observable on the model’s parameter trajectory: for each checkpoint and layer, we train a sparse autoencoder and compare feature activation profiles across the resulting layer-by-epoch grid. Each resulting heatmap is a feature-level readout of the training trajectory: it does not track movement in parameter space directly, but instead tracks how that movement manifests as the emergence, persistence, disappearance, or displacement of features.

This positions SAE-based feature dynamics as a feature-space counterpart to weight dynamics. Weight-space analyses ask where optimization moves in the loss landscape and what geometric structure the trajectory encounters; our analysis asks what interpretable representational structure is built, preserved, or reorganized along that same trajectory. The two views therefore operate at different levels of description.
%: parameter-space work characterizes the optimizer’s path through weight-space, whereas our method characterizes the feature-level consequences of that path across layers and training time.

%Our method is also related to representation-similarity analyses such as SVCCA \citep{raghu2017svcca} and CKA \citep{kornblith2019similarity}, which compare whole-layer representations across training time, architectures, or random seeds. Our method differs in granularity: instead of asking whether two layers have similar representational subspaces overall, we ask whether a specific sparse, interpretable feature has a functional counterpart at another layer and checkpoint. This places the work in the emerging SAE-based interpretability literature, where sparse autoencoders are used to recover monosemantic or more interpretable features from neural activations, and is especially close in spirit to recent work on tracking SAE features through LLM training.

In conclusion, we introduced an epoch-by-layer framework for tracking visual-feature migration in ViTs during training using SAE-based features. Using our framework, we found that features migrate mainly towards earlier layers and migration is concentrated earlier in training. In addition, we found that deeper-layer features stabilize earlier and more strongly, and that average active lifetime is longer for features in early layers. Using feature-type automation, we noticed that feature type does change over training time: early checkpoints have fewer class-specific features. The broad migration patterns are qualitatively similar across ImageNet-1k and Mixed 10, although this comparison should not be interpreted as isolating dataset scale alone, since the datasets also differ in label granularity and test-set size.% Our work opens the door to further similar explorations. 

\section{Limitations and Future Work}
This work opens a new avenue of tracking feature formation and migration by defining hypothesis-relevant metrics. It remains for future investigation to test out different similarity metrics for tracking feature evolution -- Spearman-based feature alignment can be sensitive to ties for zero activations, so, alternatives such as Top-\(K\) overlap should be explored. Currently, our SAE approach does not guarantee finding a feature across other epoch-layer pairs; this remains future work to address problematic feature discovery.

Our methodology has several limitations that suggest directions for additional future work. %First, multi-seed experiments, more datasets and larger architectures should be explored to make stronger conclusions.
%First, it is not directly applicable to architectures whose layers have different representation dimensions; adapting the proposed similarity and trajectory metrics to such settings remains an open direction.
First, since the pattern classification is to some extent subjective, then the definitions of evolution tracking metrics may need further grounding. Second, the analysis is restricted to linearly decodable features. This restriction is nontrivial, since \citet{fel2024understanding} provide evidence that not all features need to be linearly decodable throughout the network. Third, we study only supervised training. Other training regimes remain unexplored, including self-supervised objectives such as ViT-Tiny MAE \citep{he2022masked}, which can be trained from scratch with a reasonable computational budget. In addition, future work could study how feature trajectories change during fine-tuning. Evaluating the proposed methodology across multiple seeds, additional datasets, architectures, normalization schemes, and optimizers would help determine whether the observed feature-evolution and migration patterns generalize beyond the present setting.

We note that conducting a similar analysis for human-interpretable features (e.g., using linear probes with per-image attributes) would further enrich understanding of feature evolution, but is beyond this work's scope and remains a direction for future work.

A further open issue is whether feature evolution differs between the CLS and patch streams. Our methodology can be easily adapted for patch tokens. ViT analyses have shown that patch tokens retain spatial information through most layers, whereas the final layer behaves more like token mixing or learned pooling \citep{ghiasi2022vision}. In \citep{khajuria2025interpreting}, it has been shown that in multi-object images, the CLS token decodes the primary object well, but decoding of secondary objects is worse than with object-specific patch tokens. This implies the effect of the downstream task on the representation of CLS token. These examples make CLS-stream versus patch-stream feature evolution a substantive future research question for obtaining a more complete picture.
%The limitations of this work are mainly the fuel for further investigations into the realm of feature evolution.

\begin{acks}
This work was supported by the Estonian Research Council grant PRG1604, and by the Estonian Center of Excellence in Artificial Intelligence (EXAI), funded by the Estonian Ministry of Education and Research grant TK213.
\end{acks}

\section{GenAI Usage Disclosure}
%We used AI for various purposes as a helper: for relevant literature search, text refinement, code writing.
During the preparation of this manuscript, the authors utilized Generative AI technologies to assist with identifying relevant literature, improving the readability of the text, and developing code. After using these tools, the authors reviewed and edited the resulting content and take full responsibility for the presented manuscript.


\bibliographystyle{ACM-Reference-Format}
\balance
\bibliography{bibliography}
\clearpage

\appendix

\section{Appendix}\label{appendix:all}
Here we present additional results and hyperparameters that could not fit in the main text.

\subsection{The Number of Features} 
\begin{figure}[!hb]
    \centering

    \begin{subfigure}{0.48\textwidth}
        \centering
        \includegraphics[width=\linewidth]{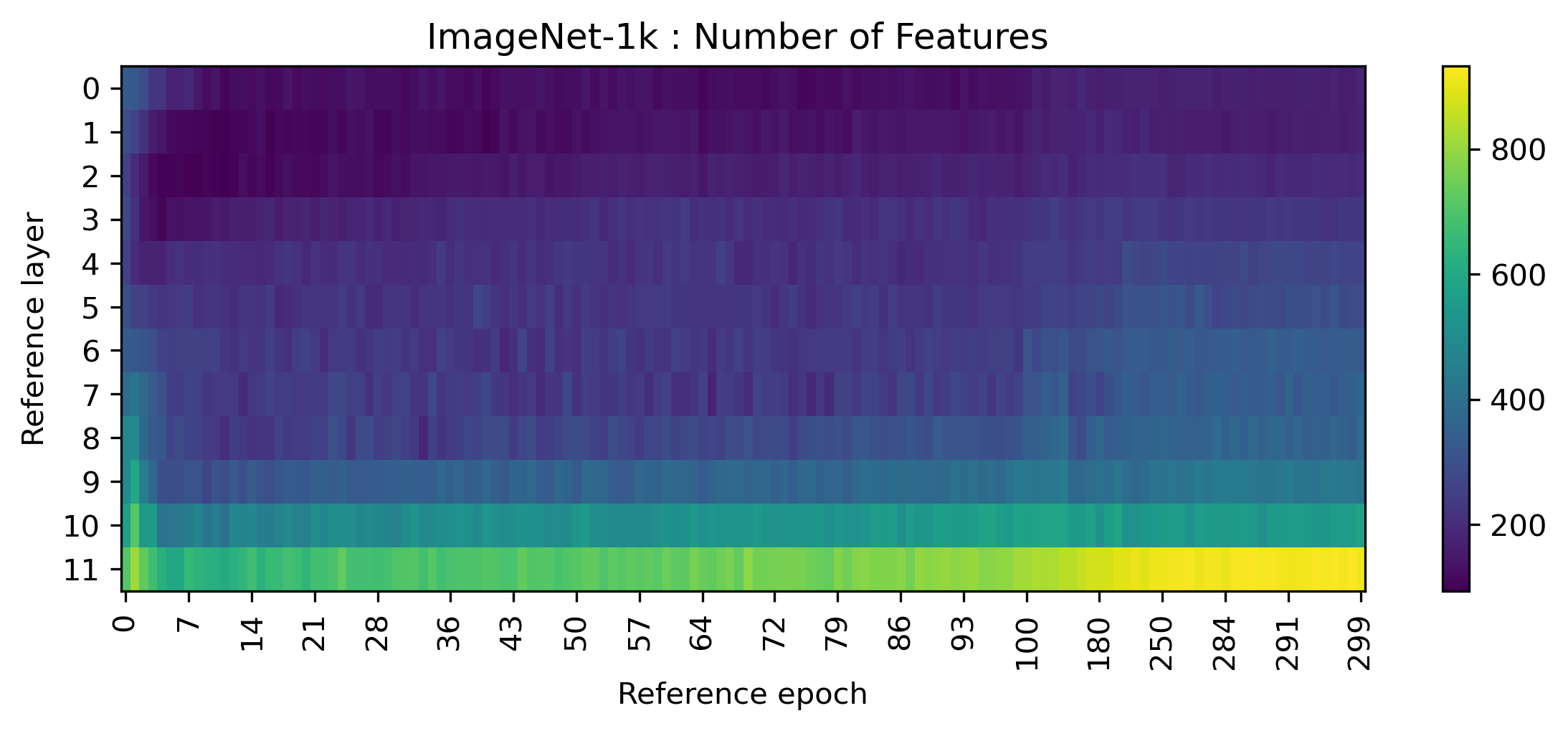}
        \label{fig:n_concepts_8}
    \end{subfigure}
    \hfill
    \begin{subfigure}{0.48\textwidth}
        \centering
        \includegraphics[width=\linewidth]{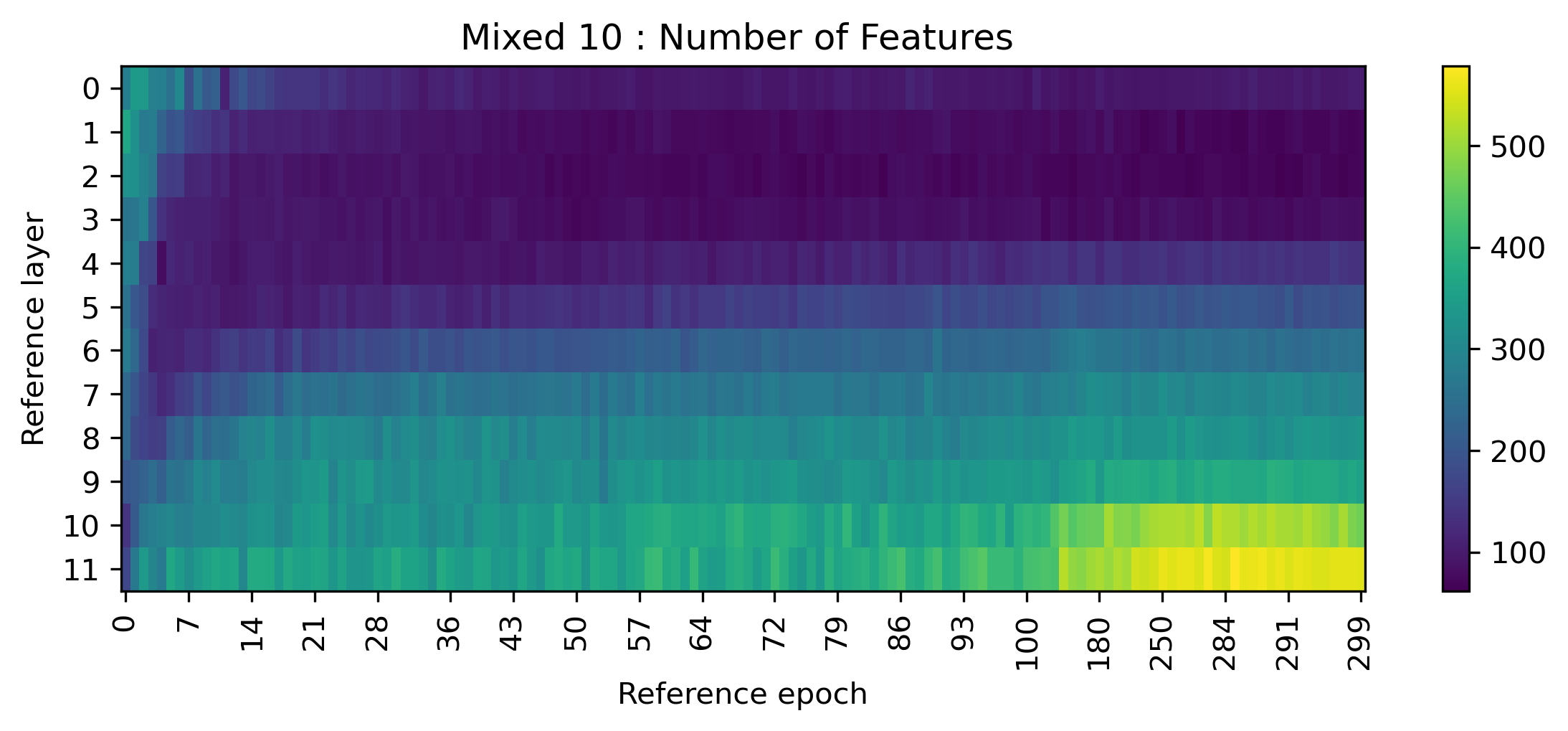}
        \label{fig:n_concepts_7}
    \end{subfigure}

    \caption{Number of features found by SAE.}
    \label{fig:n_concepts}
\end{figure}

In Figure \ref{fig:n_concepts} we report the number of SAE features that activate on more than 0.1\% of test samples i.e., the number of features discovered by the SAE in each checkpoint, providing a cue of representational complexity throughout the learning process. Otherwise the feature is considered dead (number of dead features is therefore $960-$"Number of features").

\subsection{Mixed 10 Additional Results}
\begin{figure}[!hb]
  \centering
  \includegraphics[width=\linewidth]{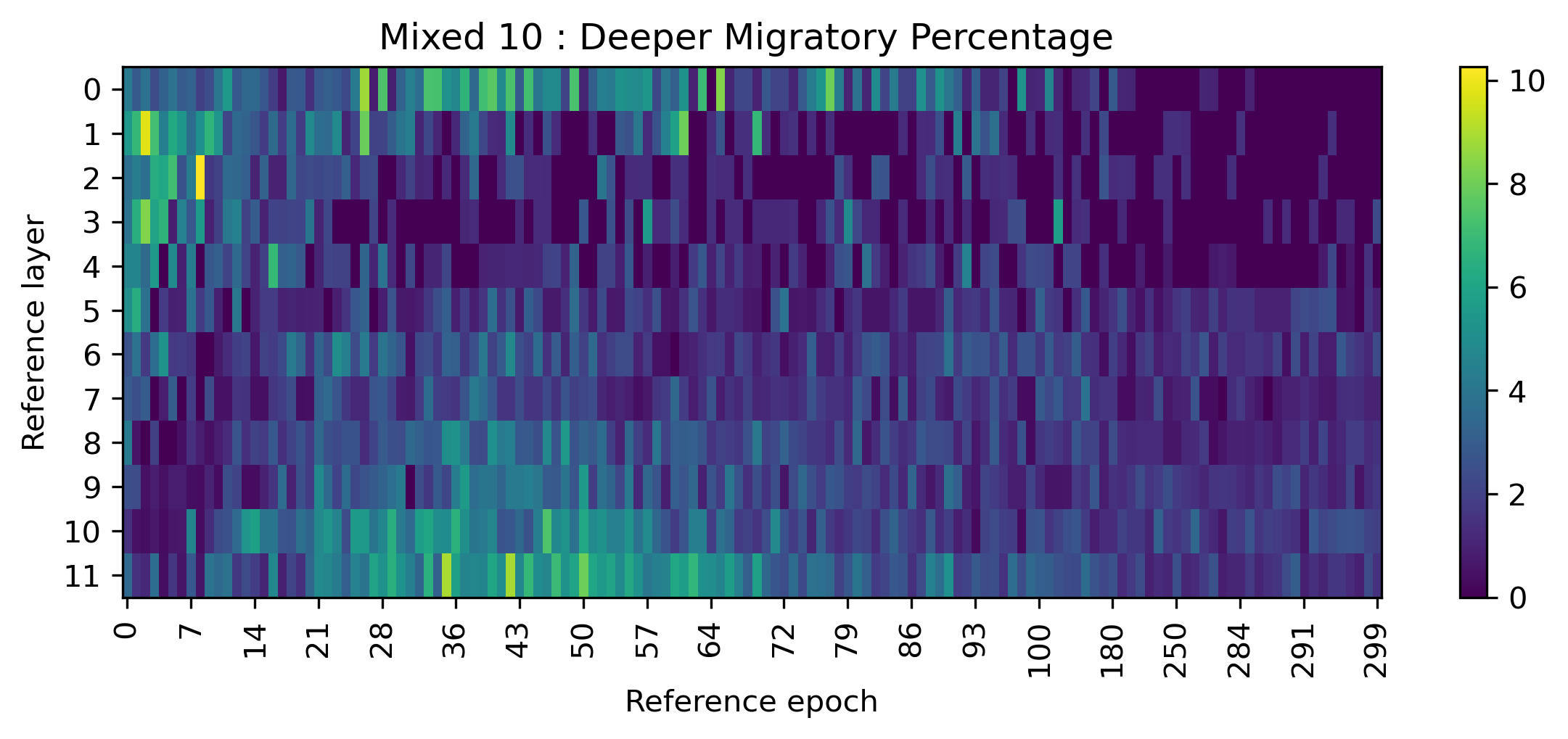}
  \caption{Percentage of features migrating towards deeper layers on Mixed 10.}\label{fig:deep_migratory_7}
\end{figure}

\begin{figure}[!hb]
  \centering
  \includegraphics[width=\linewidth]{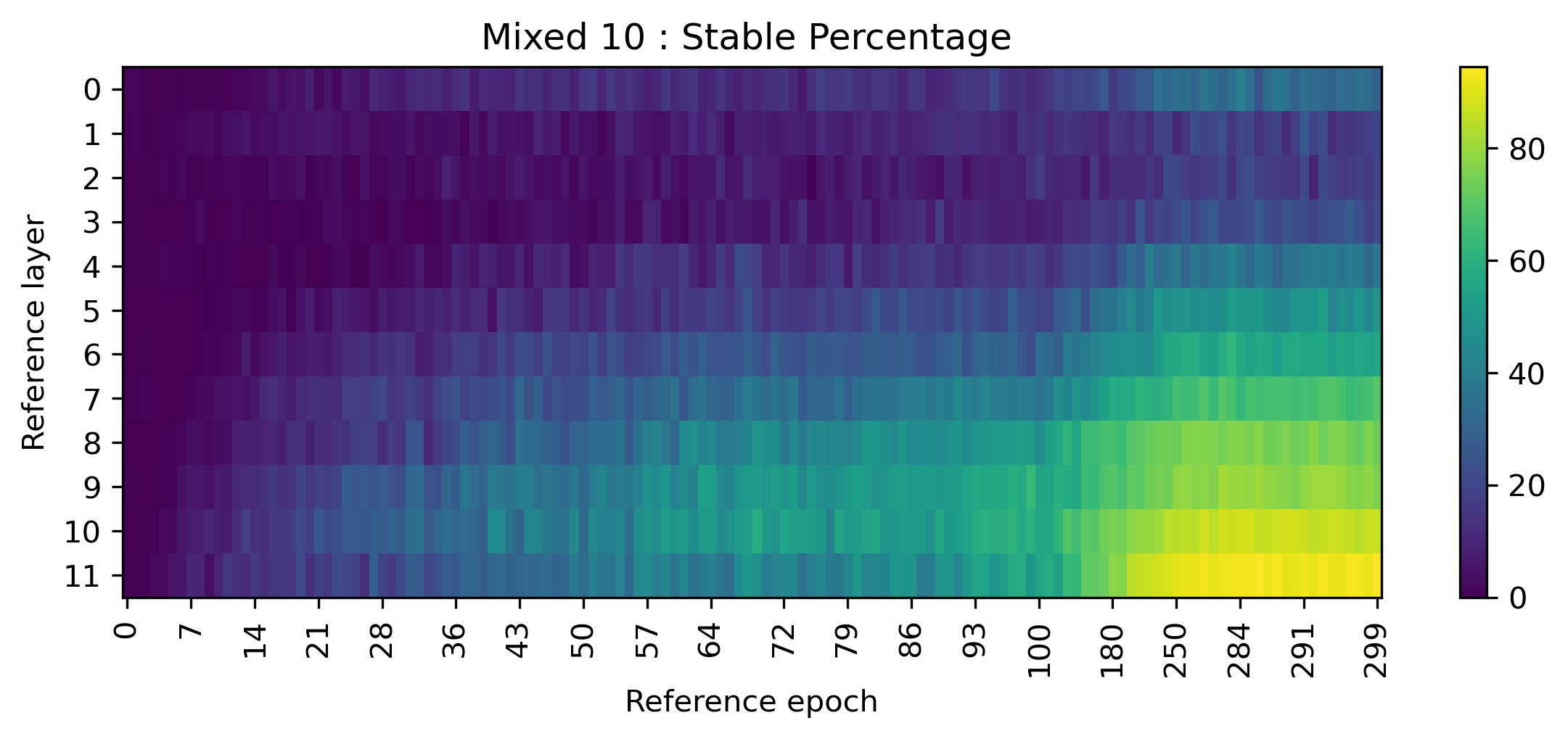}
  \caption{Percentage of stable features on Mixed 10.}\label{fig:stable7}
\end{figure}

\begin{figure}[!hb]
  \centering
  \includegraphics[width=\linewidth]{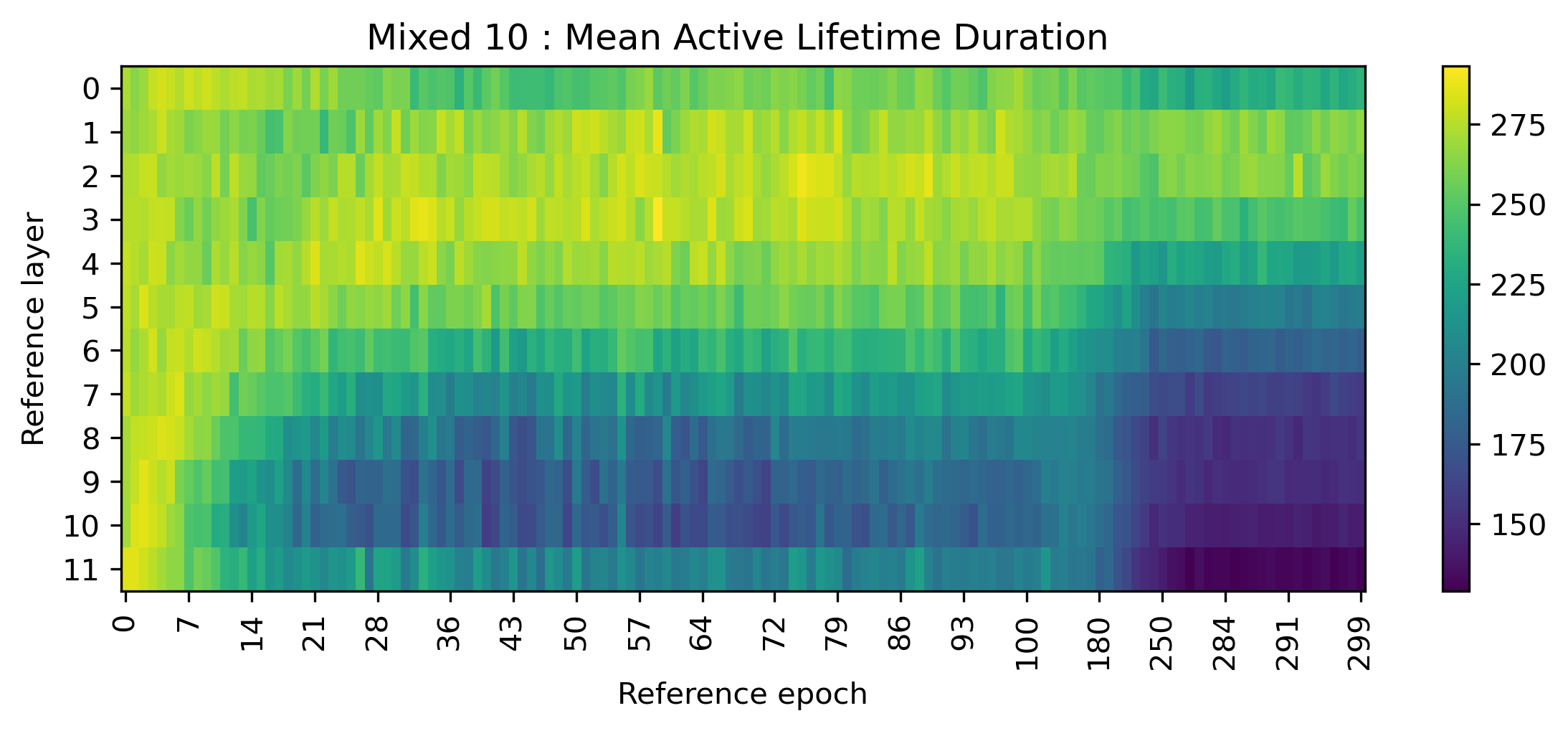}
  \caption{ Mean active lifetime of features, given in number of epochs on Mixed 10.}\label{fig:mean_active_lifetime_7}
\end{figure}

\begin{figure}[!hb]
  \centering
  \includegraphics[width=\linewidth]{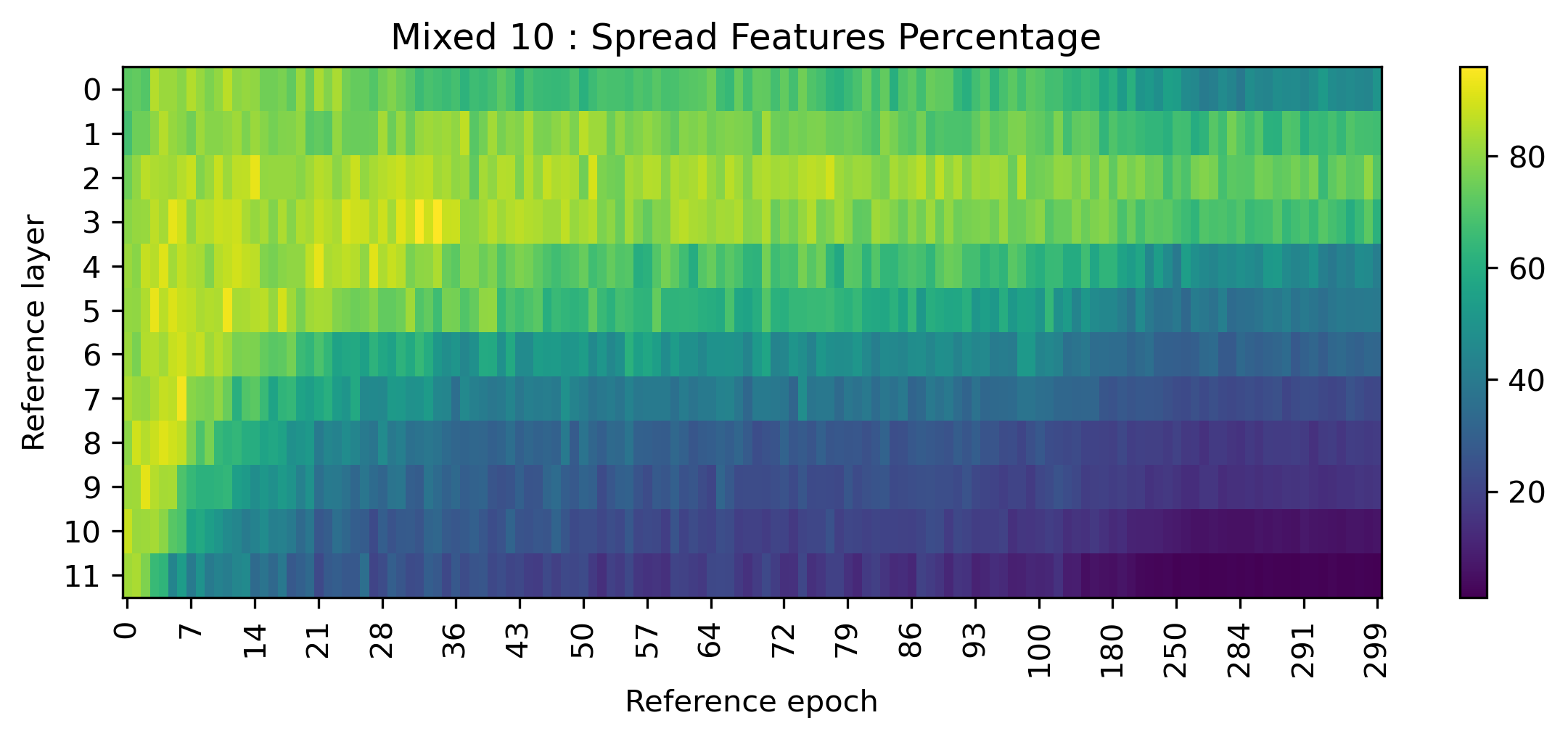}
  \caption{Percentage of features that are widely spread across layers on Mixed 10.}\label{fig:spread_7}
\end{figure}

\begin{figure}[!hb]
  \centering
  \includegraphics[width=\linewidth]{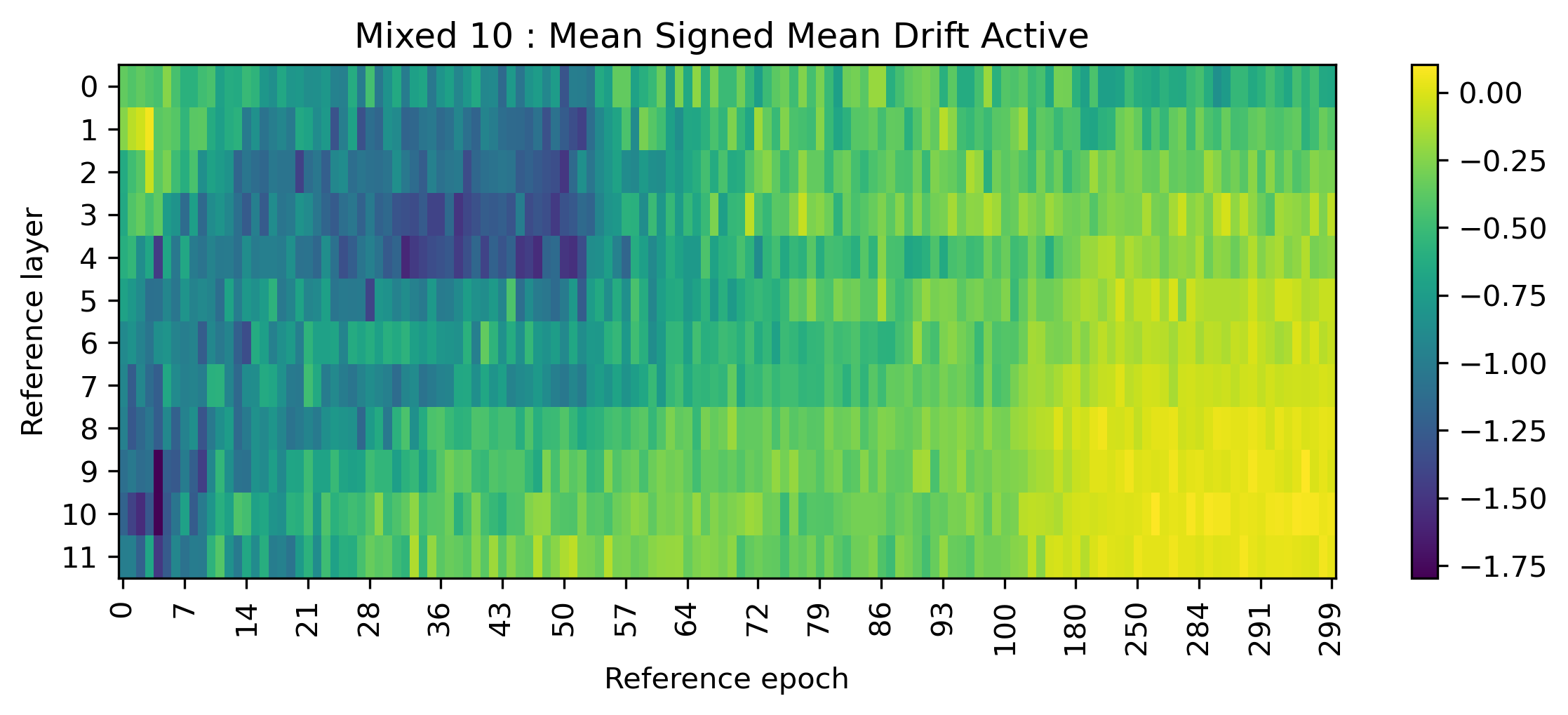}
  \caption{Mean signed active drift. For each reference epoch--layer pair, feature-level quantities are averaged across features. Average signed displacement of the layer center, in layer-index units; negative values indicate movement toward earlier layers on Mixed 10.}\label{fig:mean_signed_7}
\end{figure}

\begin{figure*}[!htpb]
  \centering
  \includegraphics[width=\linewidth]{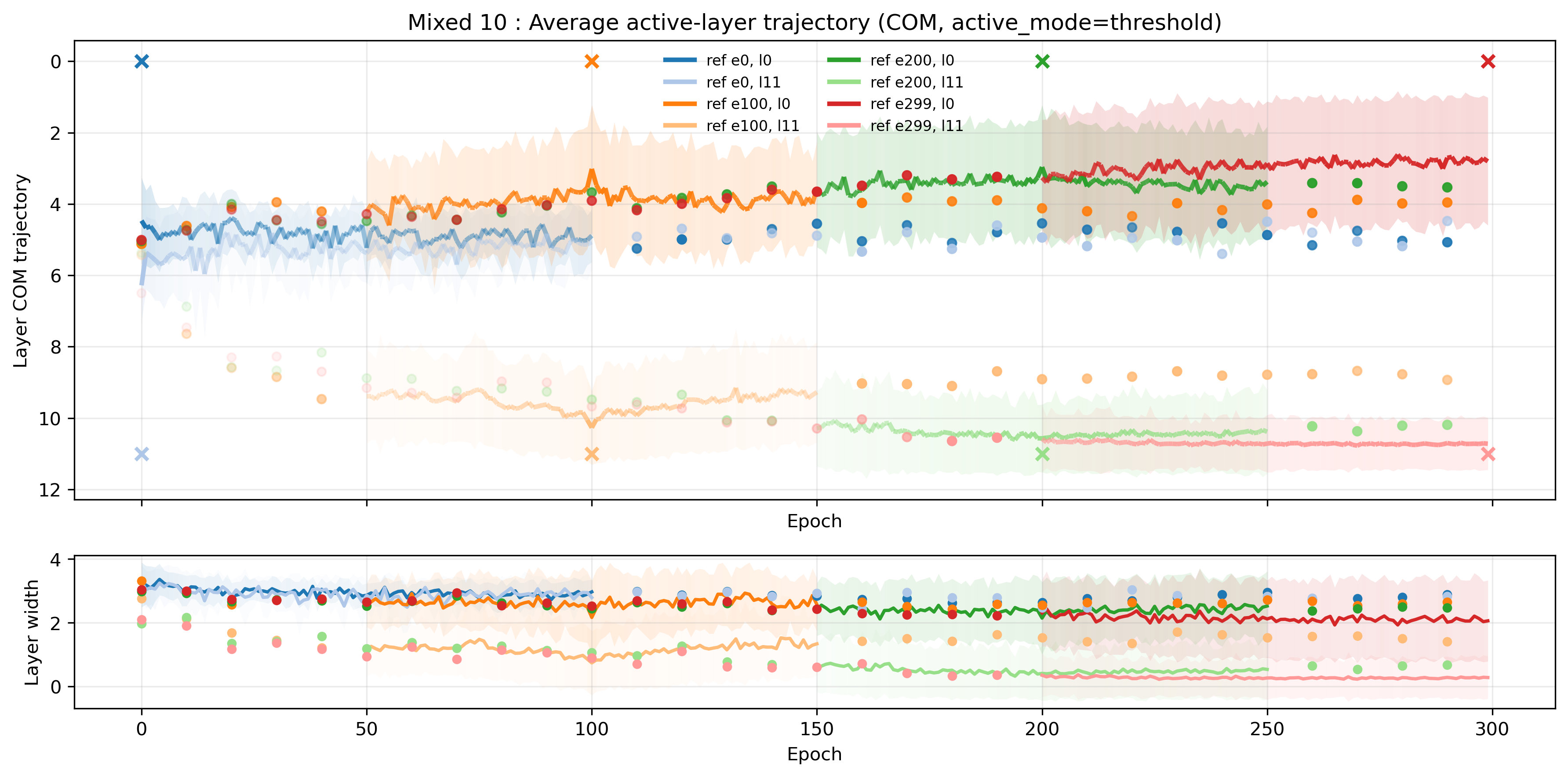}
  \caption{Layer center-of-mass (COM) $\pm$ standard deviation trajectories averaged over all the features of selected reference checkpoints (0,100,200 and 300). The high-frequency epochs are depicted as a solid line around the reference checkpoints (marked as x) and the low-frequency epochs as dots. On the figure below, the average layer width $\pm$ standard deviation is depicted. The averaging is carried out only on active epochs per feature and therefore color fading signifies the lack of active features in that area.}\label{fig:traj7}
\end{figure*}

In this section, we report the Mixed 10 counterparts that were omitted from the main paper. Deeper migratory features are depicted in Figure~\ref{fig:deep_migratory_7}, percentage of widely spread features in Figure~\ref{fig:spread_7}, mean active lifetime of features in Figure~\ref{fig:mean_active_lifetime_7}, percentage of stable features in Figure~\ref{fig:stable7}, mean signed active drift in Figure~\ref{fig:mean_signed_7} and layer center of mass trajectories in Figure~\ref{fig:traj7}.

\subsection{Feature Tracking Example}
\begin{figure*}[!htbp]
  \centering
  \includegraphics[width=\linewidth,trim={0 20.0cm 0 1cm},clip]{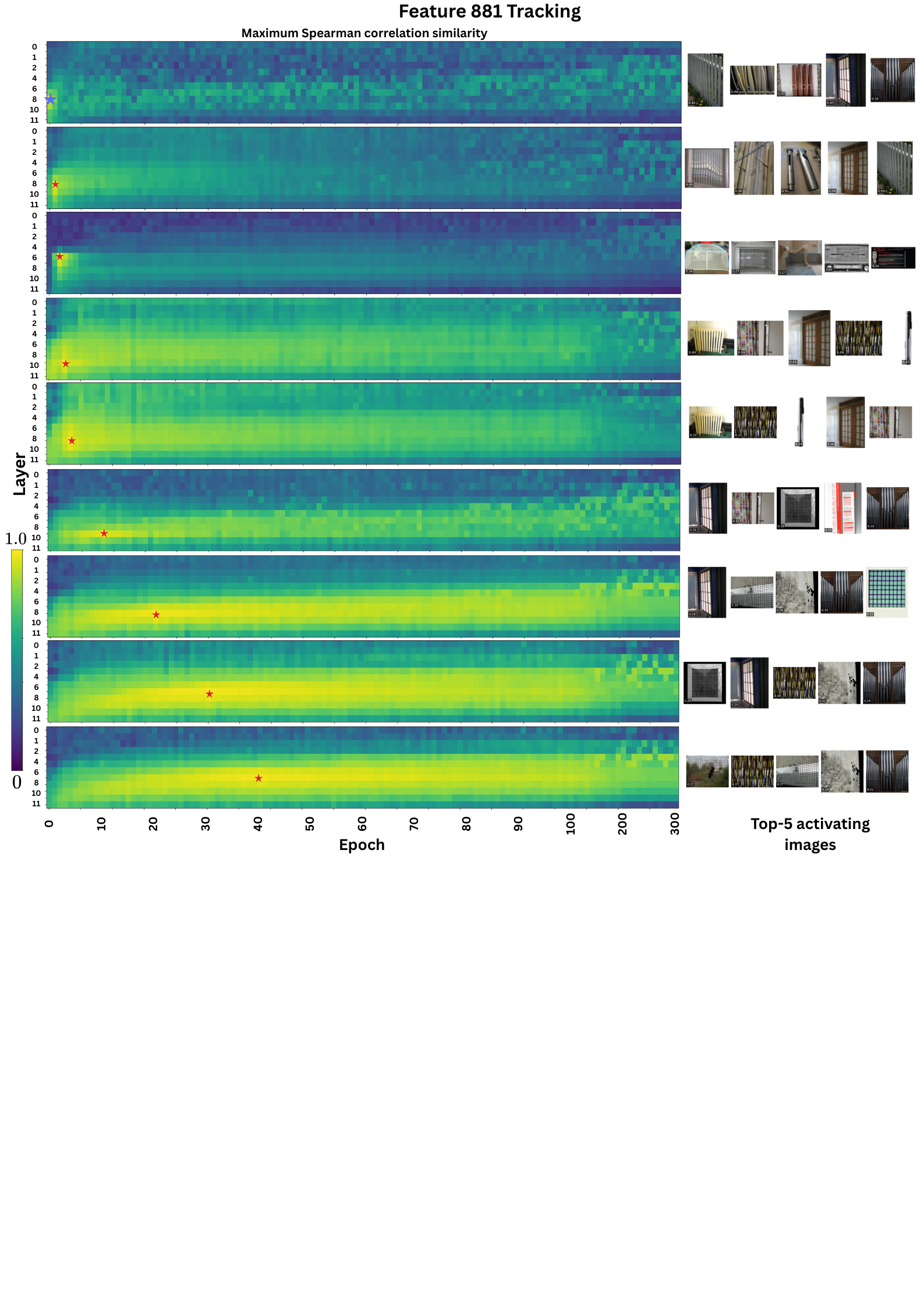}
  \caption{This figure depicts an SAE feature 881 (marked with {\color{blue}$\star$}) evolution discovered from checkpoint epoch 1, layer 8 (top figure). Figures below show the best-matching target feature at each selected checkpoint (marked with {\color{red}$\star$}) and the target feature's top-5 activating images (example image thumbnails are from the ImageNet-1k dataset).}\label{fig:tracking}
\end{figure*}

The results shown in the main part of the paper cover the aggregate perspective of feature evolution and migration; moreover, this methodology provides, as a by-product, a view into individual feature tracking. In Figure \ref{fig:tracking}, top image, we can see the evolution pattern of a specific feature starting from epoch 1 and layer 8. The images below depict the best-matching feature in the selected following epochs and their respective evolution pattern creating a clearer understanding of where the specific feature resides. This approach allows us to see the tiny scale of learning and could be used alongside annotated features. 

This also serves as a sanity check that the features matching via correlation manifest also in similar features (or at least similar top-activating images representing tube-like objects in this example).

\subsection{Computational Cost}
SAE training time and computation of epoch-by-layer views for ViT-Tiny trained on ImageNet-1k on one H200 GPU took $\approx$ 192 hours.

Due to the strict compute budget, we only explored ViT-Tiny but in future work, training larger models or including the patch-token perspective, the analysis and training pipeline should be made online as storing all the checkpoints and activations is extremely memory-heavy. Furthermore, even though we used a combination of two SAEs per checkpoint, stronger claims would require multi-seed experiments.

\subsection{SAE Loss}
\begin{figure*}[ht]
  \centering
  \includegraphics[width=\linewidth,trim={0 19.0cm 2cm 0},clip]{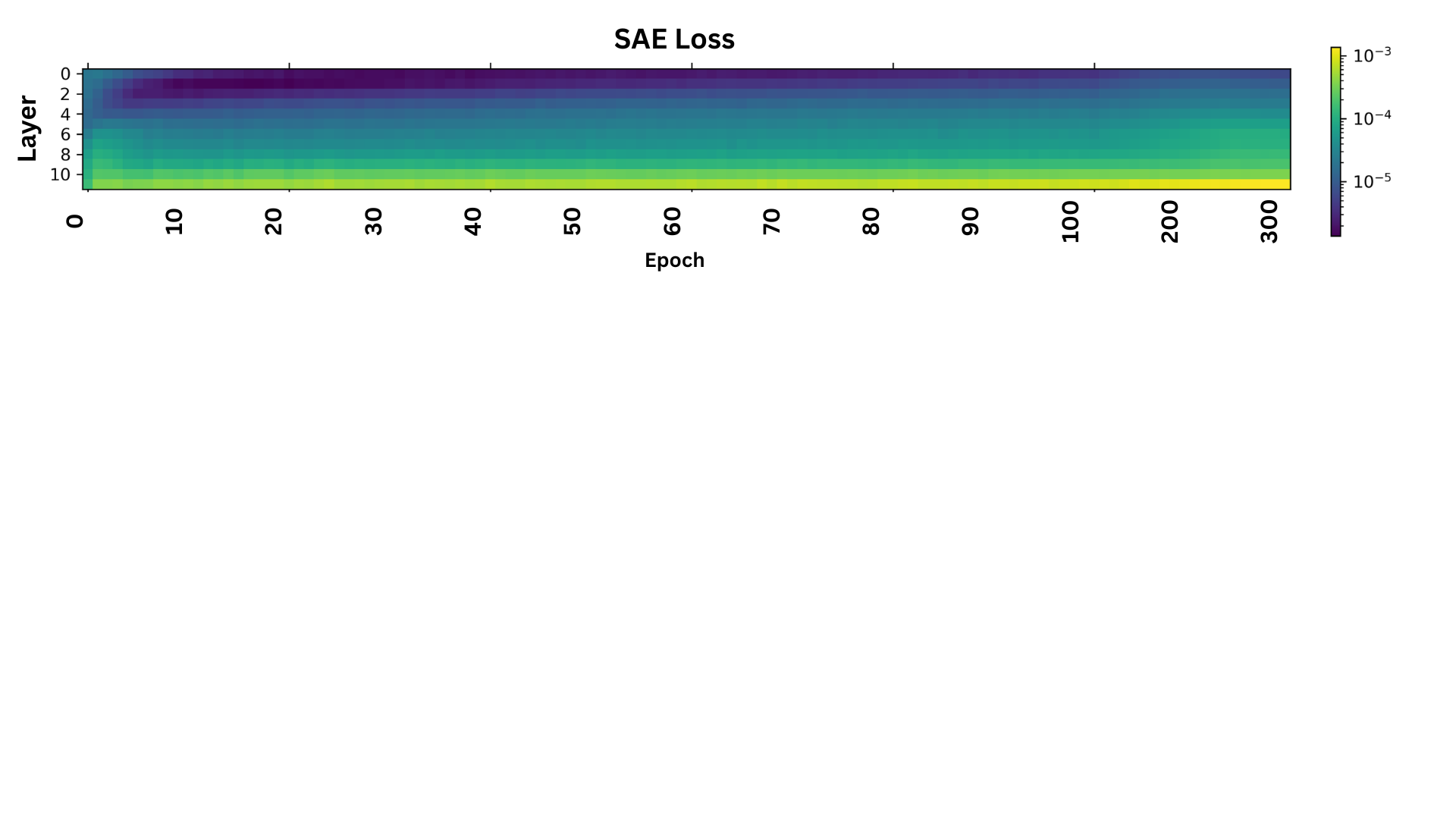}
  \caption{SAE test loss on a log-scale across time and depth on ImageNet-1k trained ViT checkpoints.}\label{fig:loss}
\end{figure*}

Since we trained all the SAEs for 150 epochs instead of training until convergence, we demonstrate the test loss of each SAE in Figure~\ref{fig:loss}. We can see that the loss is higher in the deeper layers and later in training, which correlates visually with the number-of-features analysis in Figure~\ref{fig:n_concepts}. This could be due to the complexity of representations increasing in the deeper layers and later training rendering reconstruction harder.

SAEs were trained using Adam~\citep{kingma2014adam} optimizer with a learning rate (LR) of $0.005$, without an LR scheduler or dead-feature resampling. The inputs are $\ell_2$-normalized, and each decoder dictionary vector is renormalized to unit $\ell_2$-norm after every optimizer update, thereby removing the scale ambiguity between decoder weights and feature activations \cite{gao2024scaling,bricken2023towards}.

\subsection{Comparison to CKA}
To motivate the feature-similarity view, we demonstrate the representation similarity via CKA. This is a widely used measure to compare the backbone representations $X^{\mathcal{A}} \in \mathbb{R}^{n \times d_\ell}$ and $X^{\mathcal{B}} \in \mathbb{R}^{n \times d_{\ell'}}$ on the same dataset $\mathcal{D}$. We use linear CKA \citep{kornblith2019similarity} to demonstrate the difference to our approach on column-centered representations $X^{\mathcal{A}}$ and $X^{\mathcal{B}}$ :
\begin{equation}
\mathrm{CKA}(X^{\mathcal{A}},X^{\mathcal{B}}) =\frac{\|{X^{\mathcal{A}}}^\top X^{\mathcal{B}}\|_F^2}{\|{X^{\mathcal{A}}}^\top X^{\mathcal{A}}\|_F \, \|{X^{\mathcal{B}}}^\top X^{\mathcal{B}}\|_F}.
\end{equation}

\begin{figure*}[!hb]
  \centering
  \includegraphics[width=\linewidth,trim={0 19cm 4cm 0},clip]{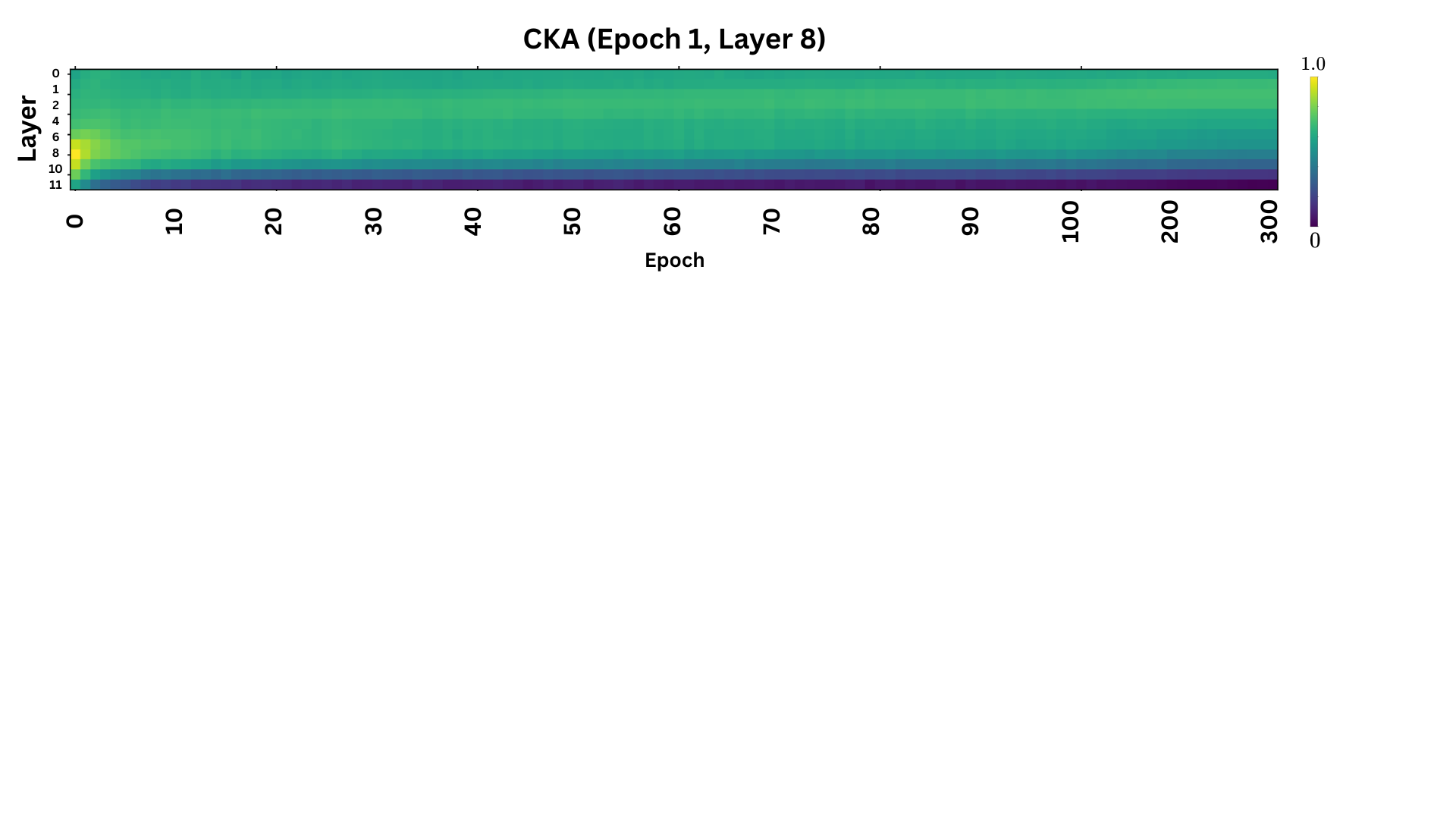}
  \caption{Linear CKA between the ImageNet-1k representation at epoch 1, layer 8 and every sampled epoch–layer pair.}\label{fig:ckaaa}
\end{figure*}

CKA provides a robust global measure of representational similarity and is invariant to orthogonal linear transformations and isotropic scaling \citep{kornblith2019similarity}. It is therefore well-suited for comparing layers and checkpoints of different dimensionalities. But CKA does not imply feature similarity: two layers may have high CKA while encoding different features, or low CKA while still supporting related features under a different parameterization \citep{vielhaben2024beyond}. In ViTs, high inter-layer similarity may also partly reflect residual pathways rather than truly shared semantics \citep{raghu2021vision, fel2024understanding}. In Figure~\ref{fig:ckaaa}, we can see an example of CKA similarity between a reference representation (epoch 1, layer 8) to every other epoch-layer pair across 12 layers and 120 epochs. 

\begin{figure*}[!htbp]
  \centering
  \includegraphics[width=\linewidth,trim={0 2.0cm 0 0},clip]{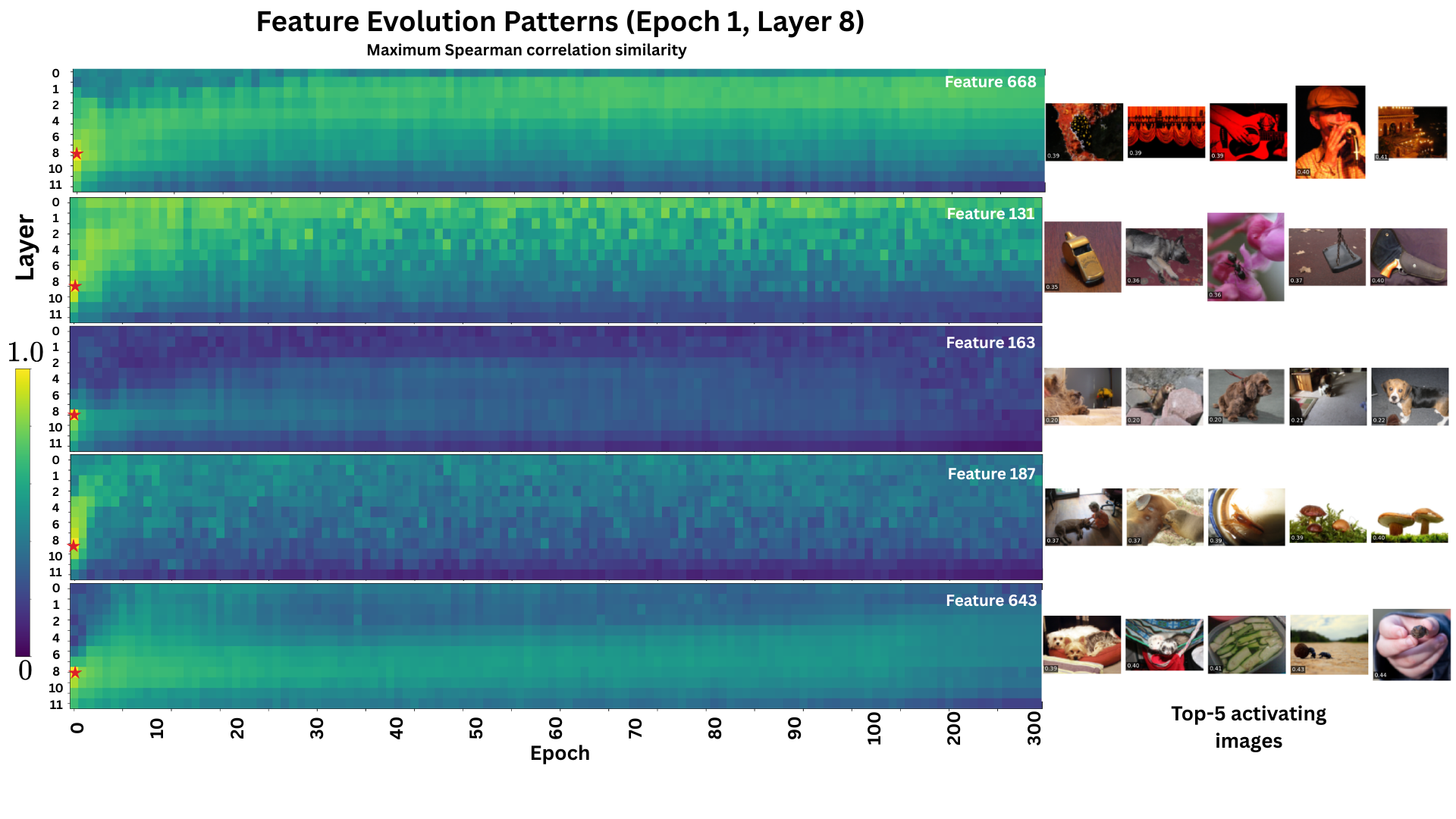}
  \caption{This figure depicts a diverse selection of SAE feature evolutions decoded from checkpoint epoch 1, layer 8 (marked with {\color{red}$\star$}) and the feature's top-5 activating images (example image thumbnails are from the ImageNet-1k dataset).}\label{fig:other}
\end{figure*}

On the other hand, in Figure~\ref{fig:other}, we can see the similarity of several chosen features from the same reference epoch 1 and layer 8 (i.e., features from the point, where it is most active or discoverable) to every other epoch-layer pair features using our method. CKA seems to capture the general trend but lacks the resolution of how the features from that reference checkpoint evolve and exhibit diverse patterns rather than a mere trend. 

\subsection{Feature Types}

\begin{figure*}
    \centering
    \includegraphics[width=\linewidth,trim={0 50 0  0},clip]{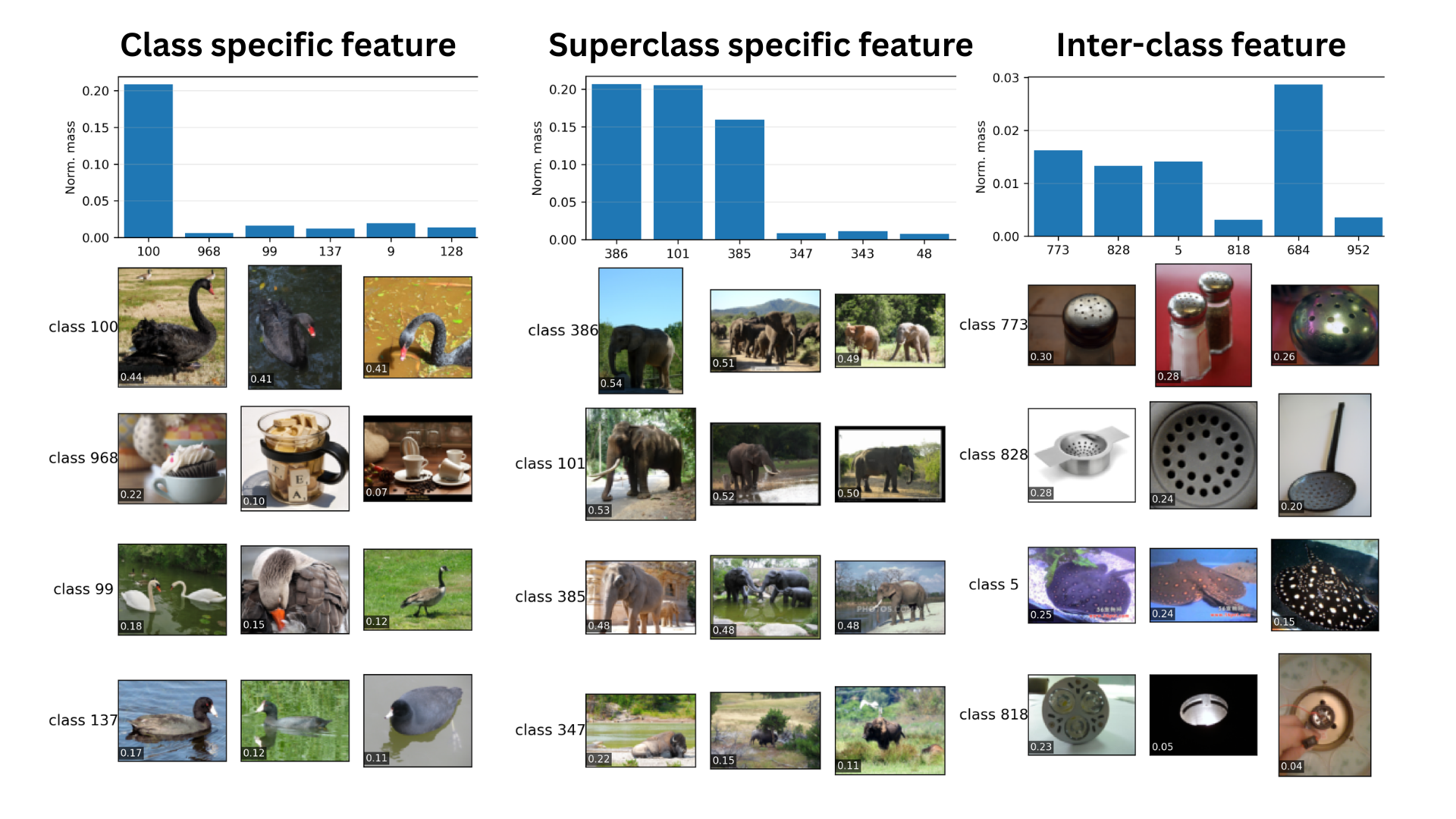} %change this to ..types_3 for high res
    \caption{Feature-type examples illustrated by maximally activating images \citep{zeiler2014visualizing}. Human-understandable SAE features are observed under 3 main categories: class-specific, superclass-specific and inter-class (Example image thumbnails are from the ImageNet-1k dataset).}
    \label{fig:feature_types_old}
\end{figure*}

Figure \ref{fig:feature_types} in the main text provides the class names for the different feature types. Figure \ref{fig:feature_types_old} additionally shows three maximally activating images for each class as illustrative examples.

\subsection{Additional Hyperparameters}
Here we report the hyperparameters for  feature evolution metrics.

For the emergence and disappearance metrics $k_{on}$ and $k_{off}$ can mean, depending on the experiment, consecutive epochs or consecutive sampled checkpoints. In our experiments, around the high-frequency window, we used only every 10th epoch, hence, $k_{on}$ and $k_{off}$ meant consecutive sampled checkpoints.
Hyperparameter values are reported in Tabel \ref{tab:feature-evolution-hyperparameters}.

\begin{table}[t]
\centering
\small
\caption{Effective feature-evolution hyperparameters.}
\label{tab:feature-evolution-hyperparameters}
\begin{tabular}{lll}
\toprule
Quantity & Value & Role \\
\midrule
$k_{\mathrm{on}}$ & $3$ & Active checkpoints for emergence \\
$k_{\mathrm{off}}$ & $5$ & Inactive checkpoints for disappearance \\
$\sigma_{\max}$ & $4$ & Maximum layer width \\
$f_{\mathrm{se}}$ & $0.05$ & Fraction of valid
active epochs at drift endpoints\\
$\delta$ & $1$ & Anchor/reference stability radius \\
$d_{\min}$ & $1$ & Strict migration threshold in layers \\
$\lambda_{\min}$ & $0.5$ & Spread-feature cutoff \\
$S_{\min}$ & $0.7$ & Minimum stability occupancy \\
$d_{\mathrm{stable}}^{\max}$ & $0.5$ & Maximum stable net drift \\
\bottomrule
\end{tabular}
\end{table}

%%\subsection{Ablation of Hyperparameters}
%%"Several design choices appear heuristic, including the max-fusion strategy and thresholds such as $\tau_c$, $|D|>1$, $|D|<0.5$, and Stability $\ge 0.7$. Some sensitivity analysis or ablation would help demonstrate that the conclusions are not dependent on these parameter choices"

\end{document}